\documentclass[letterpaper, 10 pt, conference]{ieeeconf}  % Comment this line out if you need a4paper and uncomment above line
\IEEEoverridecommandlockouts
\usepackage{CJKutf8}
\usepackage{cite}
\usepackage{balance}
\usepackage{amsmath,amssymb,amsfonts}

\usepackage{algorithm,algpseudocode}

\usepackage{comment}
\usepackage{enumerate}
\usepackage{graphicx}
\usepackage{hyperref}
\usepackage{textcomp}
\usepackage{xcolor}
\usepackage{mwe}
\usepackage{caption,subcaption}
\usepackage{svg}

\hypersetup{
    colorlinks=true,
    linkcolor=blue,
    filecolor=magenta,      
    urlcolor=cyan,
}

\newcommand{\Softbubble}{\emph{Soft-bubble} } %the space here is very important

\newcommand{\softbubble}{\emph{Soft-bubble} }

\newcommand{\forcegeometrysensor}{force/geometry sensor }

\newcommand{\forcegeometrysensors}{force/geometry sensors }

\newif\ifcomments
\newcommand{\naveen}[1]{\ifcomments {\color{blue} \textbf{Naveen}: #1} \fi}

\def\BibTeX{{\rm B\kern-.05em{\sc i\kern-.025em b}\kern-.08em
    T\kern-.1667em\lower.7ex\hbox{E}\kern-.125emX}}
    
\title{Punyo-1: Soft tactile-sensing upper-body robot for \\large object manipulation and physical human interaction}
\author{Aimee Goncalves, Naveen Kuppuswamy, Andrew Beaulieu, \\ Avinash Uttamchandani, Katherine M. Tsui, Alex Alspach \\
  Toyota Research Institute (TRI) \\
  \texttt{first.lastname@tri.global} \\
}

\begin{document}

\maketitle

% Abstract TODOs:
% - list contributions

\begin{abstract}
The manipulation of large objects and safe operation in the vicinity of humans are key capabilities of a general purpose domestic robotic assistant. 
We present the design of a soft, tactile-sensing humanoid upper-body robot and demonstrate whole-body rich-contact manipulation strategies for handling large objects. 
We demonstrate our hardware design philosophy for outfitting off-the-shelf \emph{hard} robot arms and other components with soft tactile-sensing modules, including: 
(i) low-cost, cut-resistant, contact pressure localizing coverings for the arms, 
(ii) \emph{paws} based on TRI's \Softbubble sensors for the end effectors,
and (iii) compliant force/geometry sensors for the coarse geometry sensing chest.
%We leverage the \emph{mechanical intelligence} and tactile sensing of these modules to develop and demonstrate two kinds of motion primitives: One for whole-body grasping control and one for contact-rich interaction with a human subject. 
We leverage the \emph{mechanical intelligence} and tactile sensing of these modules to develop and demonstrate motion primitives for whole-body grasping. 
We evaluate the hardware's effectiveness in achieving grasps of varying strengths over a variety of large domestic objects. 
Our results demonstrate the importance of exploiting softness and tactile sensing in contact-rich manipulation strategies, as well as a path forward for whole-body force-controlled interactions with the world. %We then discuss the implications towards designing cheaper, safer, more efficient hardware for domestic manipulation that enables more natural ways of teaching and interaction. 
\href{https://youtu.be/G8ZYgPRV5LY}{https://youtu.be/G8ZYgPRV5LY}

\end{abstract}

\section{Introduction}

\begin{figure}[h!t]
    \centering
\begin{subfigure}{0.4\textwidth}
    \centering % <-- added
    \includegraphics[width=0.95\columnwidth]{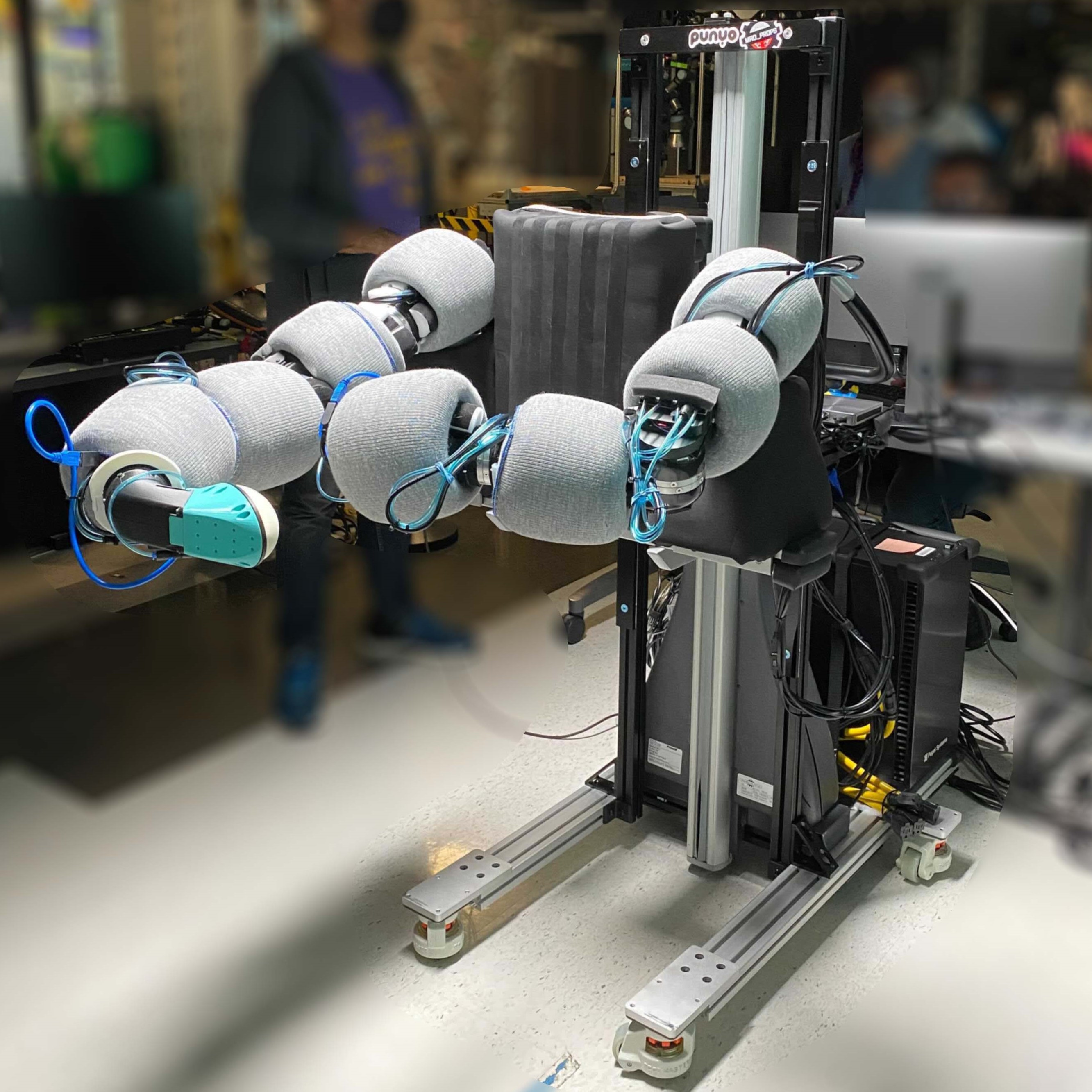} 
    \caption{}
    \label{fig:punyo-liftstick}
\end{subfigure}  
\begin{subfigure}{0.4\textwidth}
    \centering
    \vspace{-1pt}
    \includegraphics[width=0.95\columnwidth]{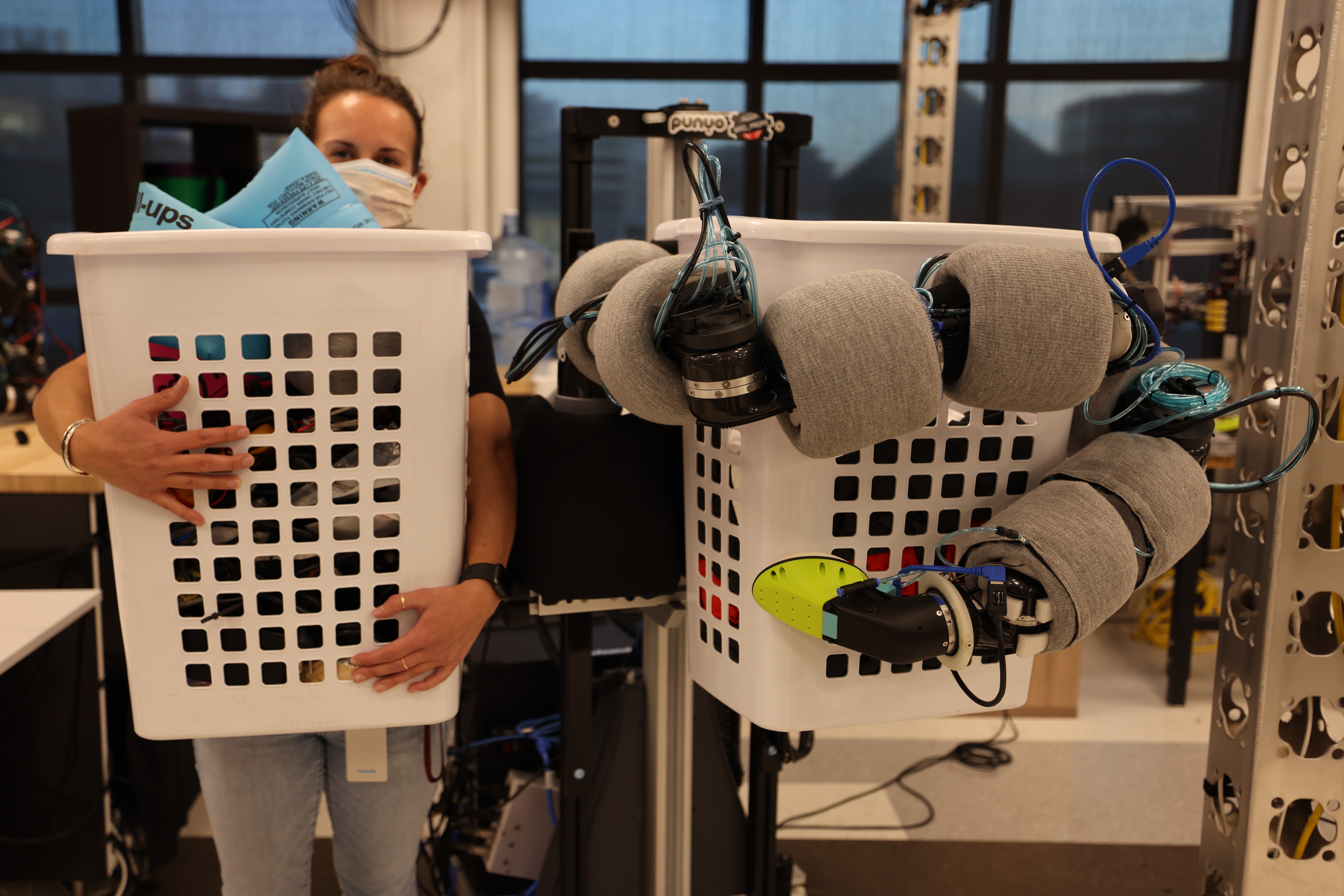}
    \caption{}
    \label{fig:side_by_side_punyo_human}
\end{subfigure} 
\vspace{-3pt}
\caption{(a) Punyo-1 upper-body humanoid robot for whole-body manipulation with soft sensing arms, end effectors, and chest. (b) Side-by-side view of human and Punyo-1 grasping a large domestic object.}
\label{fig:top-photo}
\end{figure}

\begin{figure}[t]
    \centering % <-- added
    %\begin{subfigure}{0.45\textwidth}
    \includegraphics[width=0.9\columnwidth]{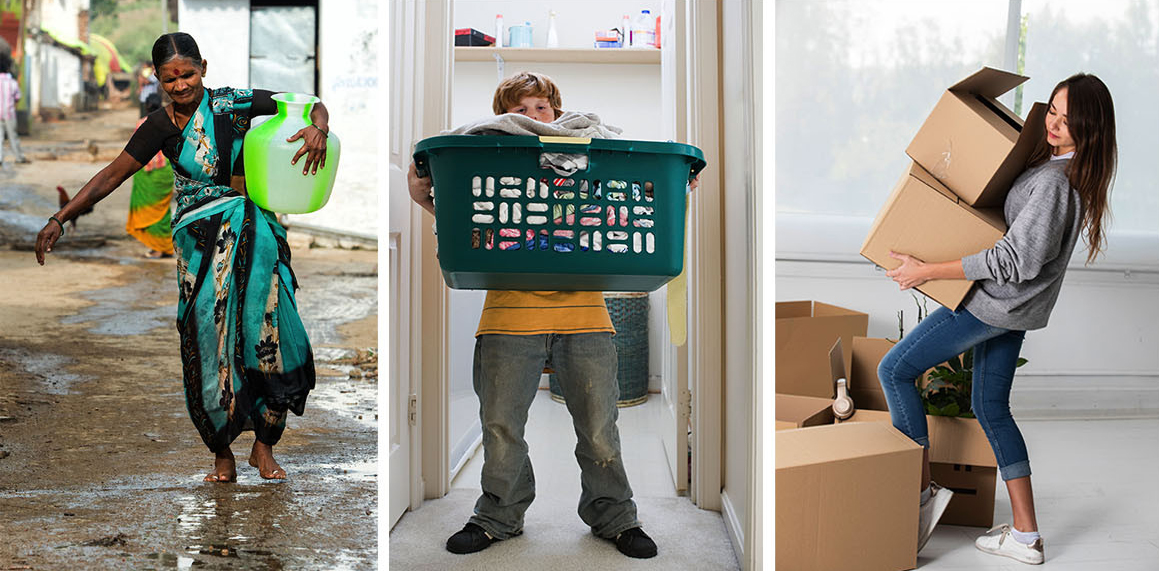}
   \caption{Humans, including older adults, often employ the whole body for household tasks.}
   \label{fig:human_contact_rich}
\end{figure}

There are many difficult problems to solve before robust robotic manipulation finds its way into our homes and daily lives.
%Most objects in our homes are designed to be held, grasped, or used with our hands, so many avenues of manipulation research rely on gripper-only grasps and interactions. 
Many avenues of manipulation research rely on gripper-only grasps and interactions, but humans often manipulate larger objects using their whole bodies with natural, contact-rich actions. Arms, chest, and other body parts are frequently used (Fig.~\ref{fig:human_contact_rich}) to carry and stabilize large or heavy objects, piles of items, or delicate items that require a gentle distribution of pressure. Our research has shown that older adults need and want assistance with large and heavy objects -- a type of assistance that may allow them to live independently longer. Co-developing effective hardware and control strategies for whole-body manipulation tasks greatly expands the capabilities of robots, especially for aiding people in their homes.

Whole-body manipulation requires innovative solutions in both hardware and control. Only a few examples of whole-body manipulation 
%control strategies 
have been reported (see Sec.~\ref{sec:background}) and most current research focuses on the planning aspect of the problem. Control of whole-body interaction remains challenging, especially in the field of physical Human-Robot Interaction (pHRI). Recent innovations in manipulator design~\cite{Hughes2016} point towards a mechanical design philosophy of \emph{mechanical intelligence}, i.e., taking advantage of passive mechanical and material properties to solve problems~\cite{kim2013soft}. Another related development in feedback-driven manipulation control uses tactile sensing~\cite{kappassov2015tactile} to better estimate contact state, but this is typically limited to small regions like finger tips. Extending these ideas, we present a mechanical design and control strategy that explicitly leverages surface compliance, rich-contact, and tactile sensing to achieve effective whole-body grasping.

\naveen{I took a pass at our core contribution below (hard robot agnostic mechanical intelligence modules)}
In this paper, we present a design philosophy for outfitting traditional \emph{hard} robots with durable, highly compliant, tactile-sensing surfaces to enable more capable, contact-rich manipulation and minimal reliance on advanced control strategies. Punyo-1, the first in a series of bimanual humanoid robots covered in our soft, tactile-sensing modules embodies this design philosophy. The word ``punyo" (\begin{CJK}{UTF8}{min}ぷにょ\end{CJK}) is a Japanese word used to describe something that is \emph{chubby and cute}, yet \emph{resilient}. 
We also present a set of grasping experiments that exemplify the advantages of compliance and tactile-sensing for whole-body grasping and lifting, object manipulation, and contact-rich pHRI.

%This paper is organized as follows; first we present a background on whole-body manipulation and upper-torso humanoid robots in Section~\ref{sec:background}. This is followed by a summary of our user needs finding research in Sec.~\ref{sec:userneeds}. We describe our mechanical design in Sec.~\ref{sec:mechanical} and our control approach in Sec.~\ref{sec:planning_and_control}. We describe several experiments in both single and dual-arm versions of our platform in order to demonstrate their efficacy in whole body manipulation and rich tactile interaction in Sec.~\ref{sec:results} which is finally followed by a discussion of future work in Sec.~\ref{sec:futurework}.

\section{Background}
\label{sec:background}
In this section, we draw motivation for domestic whole-body manipulation based on user needs, we survey robotic whole-body manipulation, and we overview the concepts of mechanical intelligence, compliance, and tactile sensing. 

\subsection{Domestic user needs}
\label{sec:userneeds}
% This is a very rough number / magnitude
% interviews and focus groups: 59+119+108+10=296
% surveys: 460+1034+97+37=1628 
Since 2016, we have observed and interviewed almost 300 older adults (age 65+) and surveyed over 1,600 in the United States and Japan (\emph{unpublished results}). 
A dominant theme is that older adults have difficulty lifting and carrying large, bulky, and/or heavy objects;
``heavy'' can be quantified as 10~lbs/5~kg in the PROMIS~\cite{cella2007patient} 
% PROMIS_Bank_v2.0_-_Physical_Function_-_11-29-2016: "Are you able to carry a heavy object (over 10 pounds / 5 kg)?"
or as ``groceries'' in the SF-36 Physical Function~\cite{brazier1992validating}.
% SF-36 PF: "carrying groceries" 
%
As people age, they experience a decrease in their mobility and stability, which are key in lifting and carrying these objects.  
Older adults are reluctant to ask for or accept help from a person as relying on someone else (e.g., familial caregiver or home health aide) can decrease their sense of independence. Some older adults even discontinue the use or purchase of such objects, which maintains their independence with a slightly reduced quality of life.   
Through our user needs finding research, we have found that some older adults are open to the idea of robotic technology physically assisting them,
especially in Japan where the ratio of older adults to young adults is 1:2~\cite{age-dependency-ratio}. %, particularly in Japan. %% I'm not sure if I want to go back to emphasizing different cultural attitudes; might be overkill for this paper?
% https://data.worldbank.org/indicator/SP.POP.DPND.OL?end=2020&locations=JP&start=1960&view=chart

\subsection{Robotic manipulation of large objects}
The field of manipulation has primarily focused on end-effector planning and control. The need for whole-body manipulation arises out of a need to also grasp objects too large or heavy for the end effector alone. Early work in this domain~\cite{salisbury1988whole, eberman1989whole} identified kinematic requirements and force-based control methods for maintaining contact and attempted to account for target geometry in the dynamics and control~\cite{song2001dynamics}. 
In the bimanual case, the planning problem has been analyzed from a geometric perspective ~\cite{seo2012spatial} on the basis of the finite nature of pairs of geometries that can be in contact and the resulting motion constraints induced.
%A related development on grasp generation planning for multi-degree of freedom (DOF) grippers lead to the notions of caging grasps, form closure, and force closure.

In contrast with planning-oriented methods, reported whole-body control methods mostly depend on joint-torque measurements and treat it as a Cartesian impedance control problem~\cite{florek2014humanoid} or explicitly aim to regulate the contact forces ~\cite{lin2018projected}. These methods rely on point-contact assumptions and derive model-based control laws to regulate the state. More recent works have also explored utilizing reinforcement learning to synthesize policies for whole-arm manipulation~\cite{yuan2019reinforcement}. 

%  In the bimanual case, an approach for planning whole-body grasps using generalized contact was proposed for polytope geometries~\cite{seo2012spatial} and demonstrated using a PR2 robot. Machine learning methods were also deployed for solving this multi-contact planning problem using reinforcement learning techniques~\cite{yuan2019reinforcement}.  

% Florek~\cite{florek2014humanoid} proposed a Cartesian impedance control scheme for multi-arm control using a model composed of multiple virtual springs. The need to simultaneously regulate manipulator configurations and the contact-forces required to maintain the target object at a desired state inspired methods that optimize torques on the projected inverse dynamics~\cite{lin2018projected}. 

While the kinematic/planning oriented approaches do not address robustness to disturbances (e.g., unexpected contact), both planning and model-based control methods require explicit object models, and in either case, grasp-generation control is not explicitly addressed. Moreover, none of these methods adequately address the challenging nature of the contact-dynamics between robot and object. Some challenges of contact-rich whole-body manipulation are similar to that of dexterous gripper grasp planning and control~\cite{ozawa2017grasp}:
\begin{itemize}
    \item accounting for closed-kinematic chains of multiple arms making and maintaining contact with a target geometry,
    \item enumerating and maintaining rich contact (more than two independent contact regions) with the target object,
    \item accounting for the frictional and compliance properties of the contact, and 
    \item observing the state of the target object and contact area.
\end{itemize} 

While numerically intensive planning with precise control or carefully trained machine learning can yield functional whole-body grasps, the field of soft robotics \cite{kim2013soft} has shown that incorporating mechanical intelligence into hardware designs could help leapfrog many of these underlying challenges. Compliant materials on robot surfaces can alleviate the burden of active compliance on grasping controllers and offer large, high-friction contact patches for stable form and force closure. Furthermore, these large contact patches provide 
%a wide 
an opportunity for tactile sensing-based state estimation.

\subsection{Compliance and tactile sensing}
For maintaining contact and regulating object state during grasping and manipulation, it has been recognized that incorporating under-actuated mechanisms, compliance, and friction offers tremendous gains in manipulation robustness~\cite{Hughes2016}.
Effectively using compliant contacts all over the body was demonstrated by Alspach et al.~\cite{Alspach2015}, in which pressure-based feedback informed grasping of unmodeled objects. 

Inspired by these ideas, a gripper developed by Toyota Research Institute (TRI), known as the \Softbubble (or \emph{Punyo}) gripper~\cite{Kuppuswamy2020,punyotech}, incorporates highly compliant tactile-sensing fingers~\cite{Alspach2019} and not only maintains passively stable grasps but leverages multi-modal sensing to detect contact and manipuland state. The gripper consists of two \Softbubble sensors that integrate a ToF camera within an inflated elastic membrane. This design not only enables highly compliant interaction with object surfaces but through integration of a pressure sensor and a deposition of markers within the membrane facilitates sensing of contact pressure, geometry and shear forces simultaneously.  

Various tactile sensing skins have been developed for both robot grippers~\cite{kappassov2015tactile} and for the body of the robot~\cite{Kim2015, cannata2008embedded}. These potentially enable whole-body interaction with objects, humans and the environment~\cite{mittendorfer2015realizing}, and are a promising solutions for multi-contact control, as evidenced by results on monitoring contact patch geometries~\cite{kuppuswamy2019fast}, and in developing manipulation primitives~\cite{hogan2020tactile}.

Building upon these ideas, we demonstrate an approach for augmenting off-the-shelf \emph{hard} robot platforms with surface compliance and tactile sensing. We also develop a simple grasping controller that relies on compliance and tactile feedback. 
Our approach is perhaps best compared against the tactile feedback approach shown by Mittendorfer 
%and colleagues
et al.~\cite{mittendorfer2015realizing}, although their methods have only been demonstrated on a limited range of object weights and sizes. Moreover, in comparison, we demonstrate robust grasping with sensing that is much lower in spatial resolution but much more compliant, allowing for passive surface behavior to maintain the grasp. This is best exemplified by the ability of our robot to apply sufficiently large forces to lift objects that are well above the stated payload capacity of the arms.
With a strong focus on exploiting mechanical intelligence, we enable robust whole-body grasping for large, unmodeled domestic objects with comparatively simpler tactile feedback-based control strategies.

%In the next section, we present our mechanical design and discuss our soft, tactile sensing modules.

\section{Upper-Body Mechanical Design}
\label{sec:mechanical}

%To explore large object manipulation, we designed and fabricated a soft bimanual upper-body platform.
We designed and fabricated a soft, bimanual upper-body platform, depicted in Fig.~\ref{fig:top-photo}. This experimental platform is composed of off-the-shelf robot arms augmented with compliant sensors on the links, end effectors, and chest. The upper-body assembly is mounted to a vertical lift with passive wheels so that it can be moved up, down, and around for grasping experiments.

\subsection{Kinematic exploration}
\label{sec:kinematic}

The upper-body consists of two Kinova Jaco Gen2 7DOF robot arms~\cite{jacogen2}
%~\footnote{See https://www.kinovarobotics.com/en/products/gen2-robot for more information} 
mounted vertically on either side of a compliant chest. The arms and chest are mounted in-line on a rail so that the shoulder width can be adjusted. The bottom of the chest aligns with the base of the arms, extends upward 42.8~cm and is 28.5~cm wide. The convex chest extends 
8~cm 
forward beyond the central axis of each arm's first link. %by 8~cm. 

\begin{figure}[t]
    \centering
\begin{subfigure}{0.24\textwidth}
    \includegraphics[width=\linewidth]{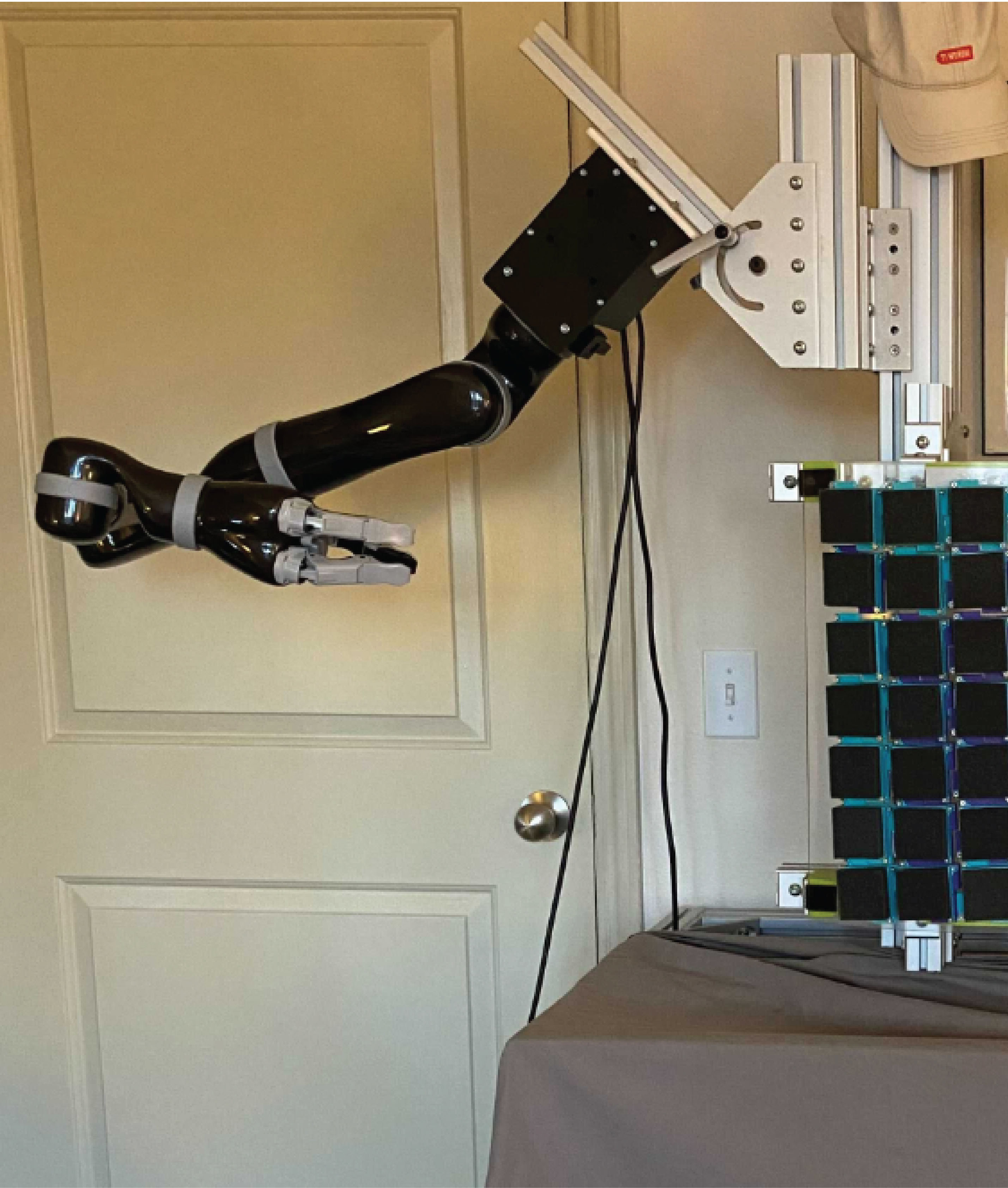}
    \caption{}
\end{subfigure}\hfil
\begin{subfigure}{0.22\textwidth}
    \includegraphics[width=\linewidth]{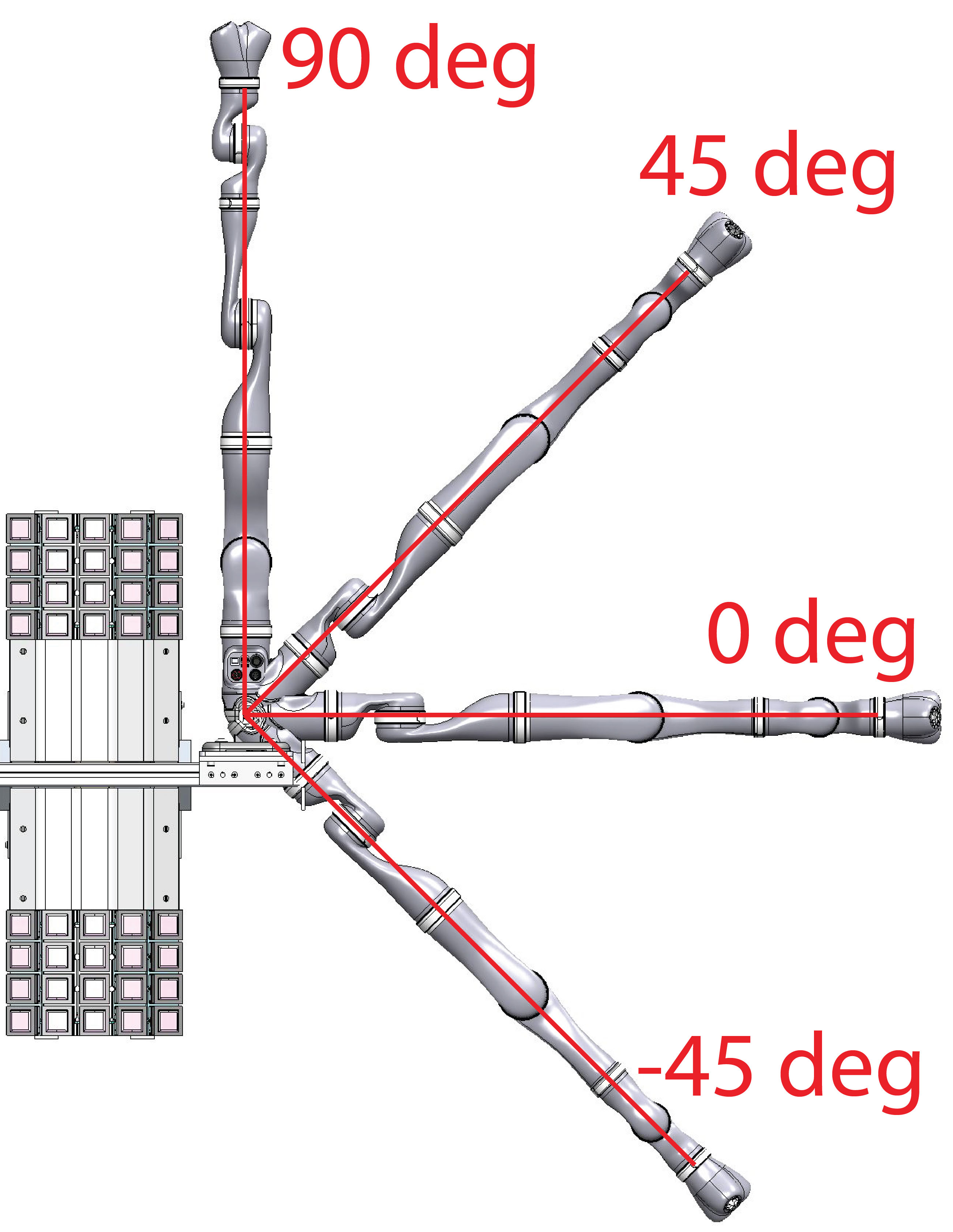}
    \caption{}
\end{subfigure}\hfil 
%\caption{ (a) Test rig used for exploration in -45 deg configuration, (b) Schematic of the tested shoulder angles with respect to chest support structure.}
\caption{ (a) Exploratory test rig set at -45 deg, (b) Tested shoulder angles with respect to chest support structure.}
\vspace{-3pt}
\label{fig:jaco-angle-CAD}
\end{figure}

The configuration of the arms and chest relative to one another, especially the shoulder angles, significantly impacts the whole-body manipulation workspace. 
To determine the most qualitatively capable starting configuration for grasping experiments, 
manipulator reachability, 
and ability to perform bimanual upper-body grasps, 
the arms were manually driven to explore grasping a set of large domestic objects (e.g., hamper, pasta pot) while varying the shoulder angle from -45$^\circ$ to 90$^\circ$ (Fig.~\ref{fig:jaco-angle-CAD}). 
To ensure stable grasps, passively compliant, non-sensing structures were added to the robot's contact surfaces (e.g., thin, high-friction rubber stickers, and off-the-shelf inflated structures like inflatable armbands). From experimentation, a shoulder angle of 90$^\circ$ provided the best reachability.

The upper body was designed to allow for adjustability for grasping experiments. The robot arms are mounted on horizontal sliders so the shoulder width can be adjusted between 45.8 and 55~cm (measured from the arms' first link central axis); for the experiments herein, we used a shoulder width of 52~cm. The chest support structure provides mounting locations for quickly changing the mechanical and sensing properties of the chest. The adaptability of the upper-body structure enables us to converge, through experimentation and adjustments, on an upper-body anatomy appropriate for manipulating a wide range of domestic objects.

%Morphological exploration: (a) Examples of passive compliance added to the robot arms surfaces; 1: inflatable armband, 2: non-slip furniture pad, 3: 3/8" thick neoprene rubber foam, 4: 1/4" thick neoprene sponge, 5: textured gripping pads (b)

\subsection{Vertical lift and passive mobility}
\label{sec:vertical}

The entire upper-body is secured to a manually operated vertical lift shown in Fig.~\ref{fig:punyo-liftstick}. The vertical lift allows for the adjustment of the upper body to a specific grasping height. The base of the chest can travel vertically from the floor to a height of 140~cm. The upper body and vertical lift assembly sit on a platform with locking, leveling passive casters. This platform also supports the control computer and other electronics. With a power and networking cable bundle running off the platform, the system can be wheeled around to transfer grasped objects from one location to another. 

The choice to build upper body hardware atop a manually operated vertical lift mechanism and a passive wheeled base is two fold: (1) The system represents a mobile manipulation platform, either wheeled or legged, without distracting development focus from the core manipulation goals being explored, and (2) the human element prevents a reliance on ground truth knowledge of the manipuland's state,  forcing an embrace of tactile-driven feedback control.
%robot-manipuland relationship forcing an embrace of tactile-driven feedback control.

\section{Compliance and tactile sensing}
\label{sec:compliance}

The three types of modules used to cover Punyo-1's arms and chest combine the benefits of highly-deformable passive compliance with the richness of tactile sensing. The individual sensing module capabilities described here are stand-alone and can be used to augment more surfaces of future Punyo revisions, as well as other robotic platforms.

\subsection{Module i: Pressure-Sensing Chambers on Arm Links} 
\begin{figure}[t]
    \centering % <-- added
    %\begin{subfigure}{0.45\textwidth}
    \includegraphics[width=0.9\columnwidth]{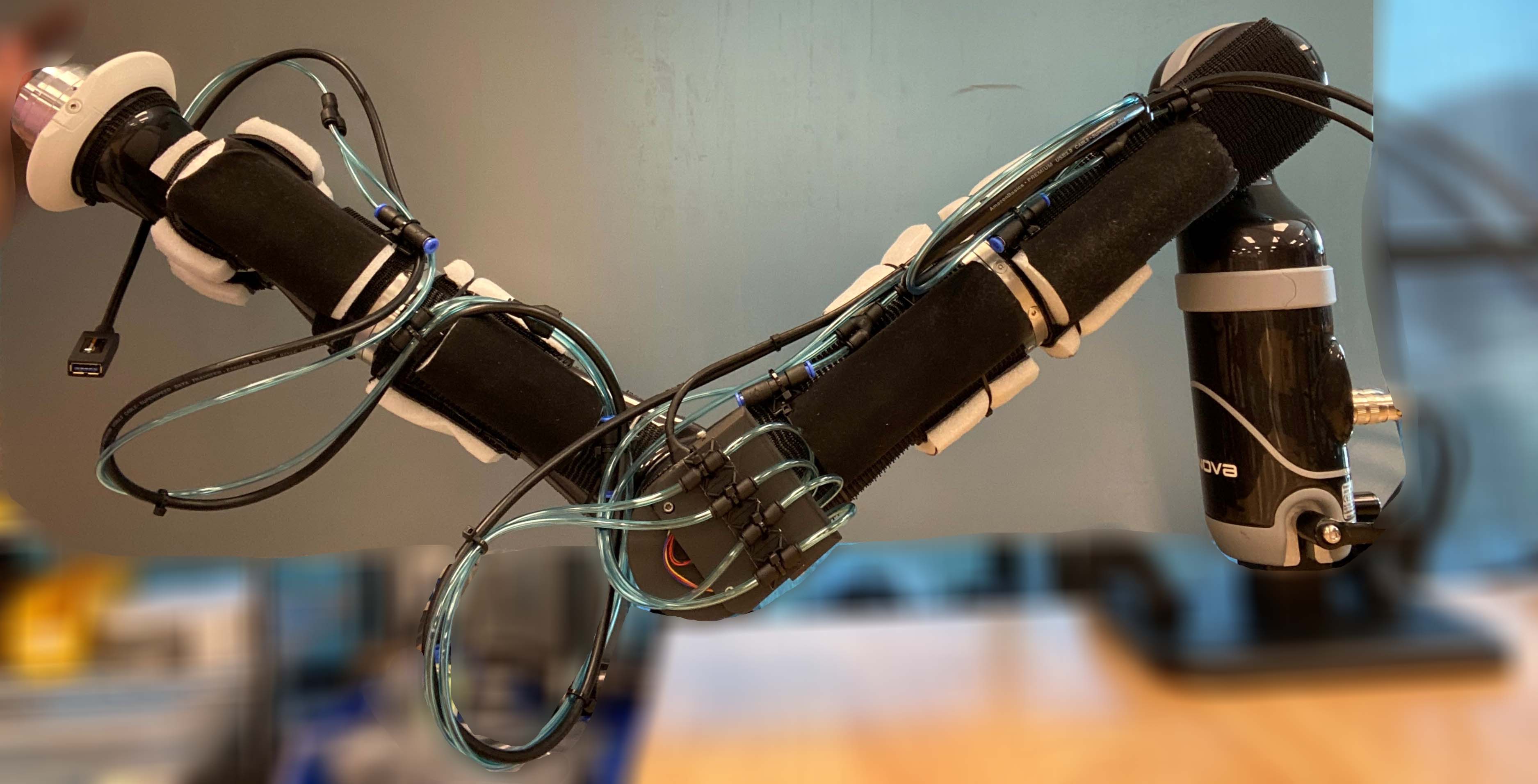}
    \caption{High-friction, anti-rotation foam structures and air-tube/cable routing for pressure-sensing chambers.}
    \label{fig:arm-tube-routing}
\end{figure}

\begin{figure}[t]
    \centering % <-- added
\begin{subfigure}{0.15\textwidth}
  \includegraphics[width=\linewidth]{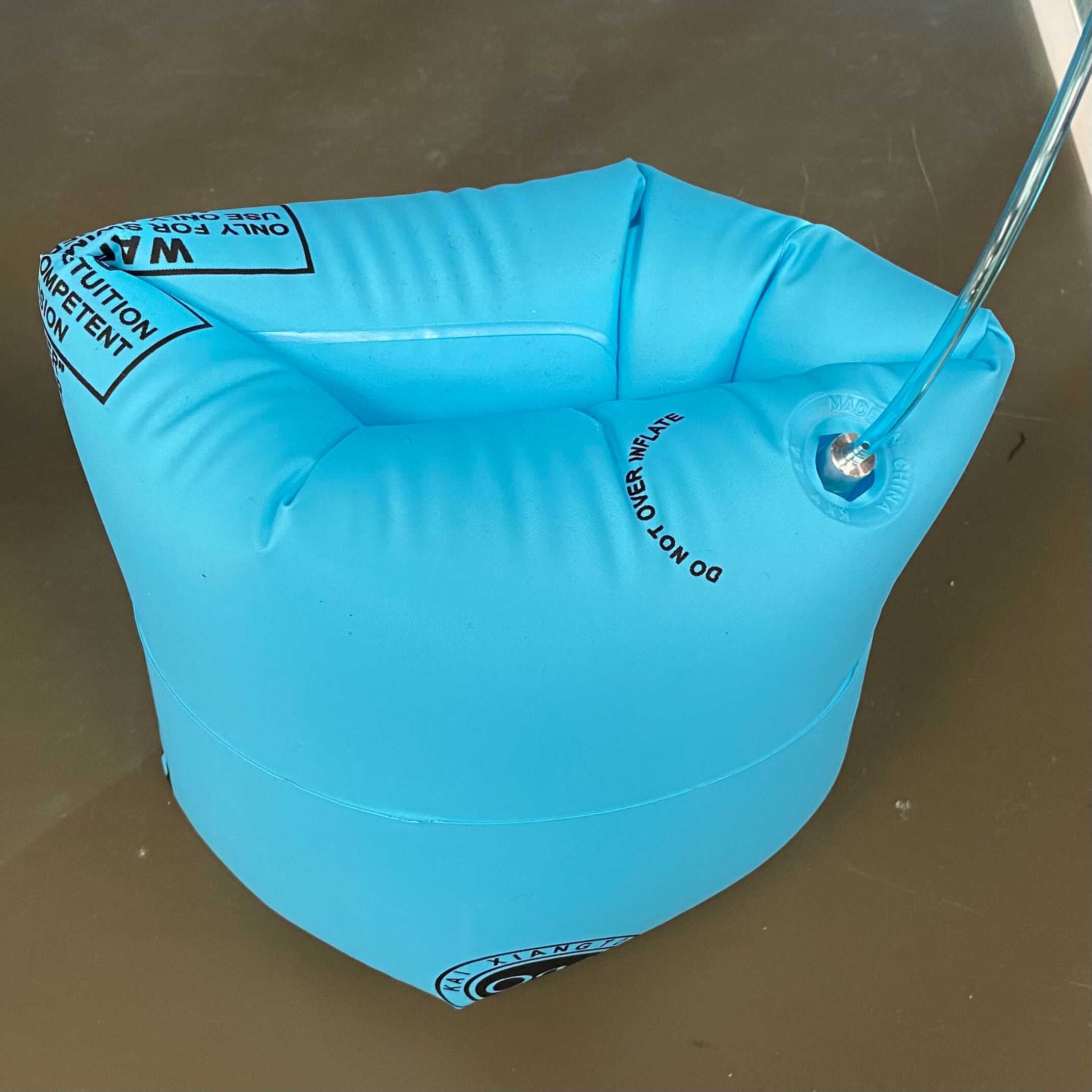}
  \caption{}
  \label{fig:floatie:plain}
\end{subfigure}\hfil % <-- added
\begin{subfigure}{0.15\textwidth}
  \includegraphics[width=\linewidth]{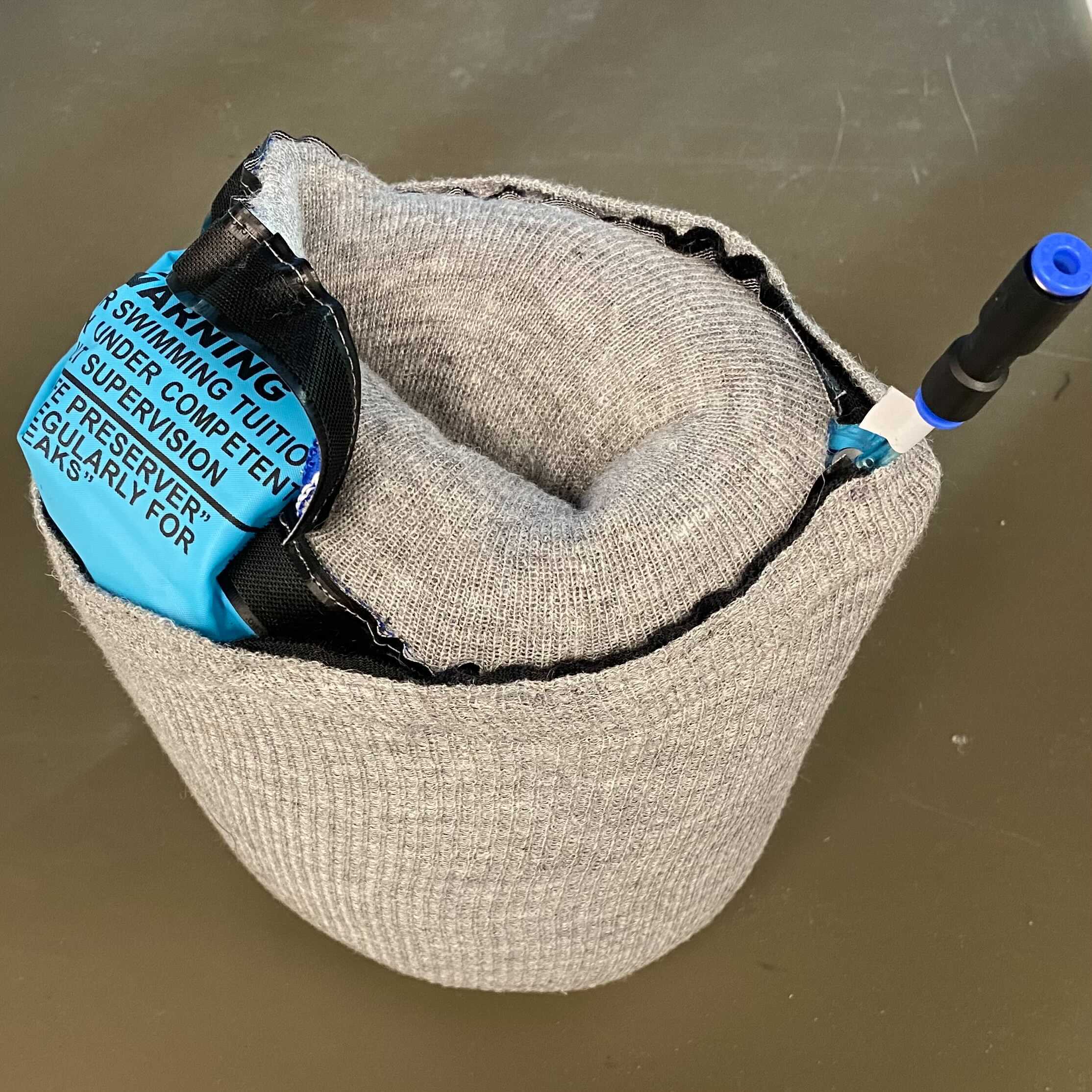}
  \caption{}
  \label{fig:floatie:halfcover}
\end{subfigure}\hfil % <-- added
\begin{subfigure}{0.15\textwidth}
  \includegraphics[width=\linewidth]{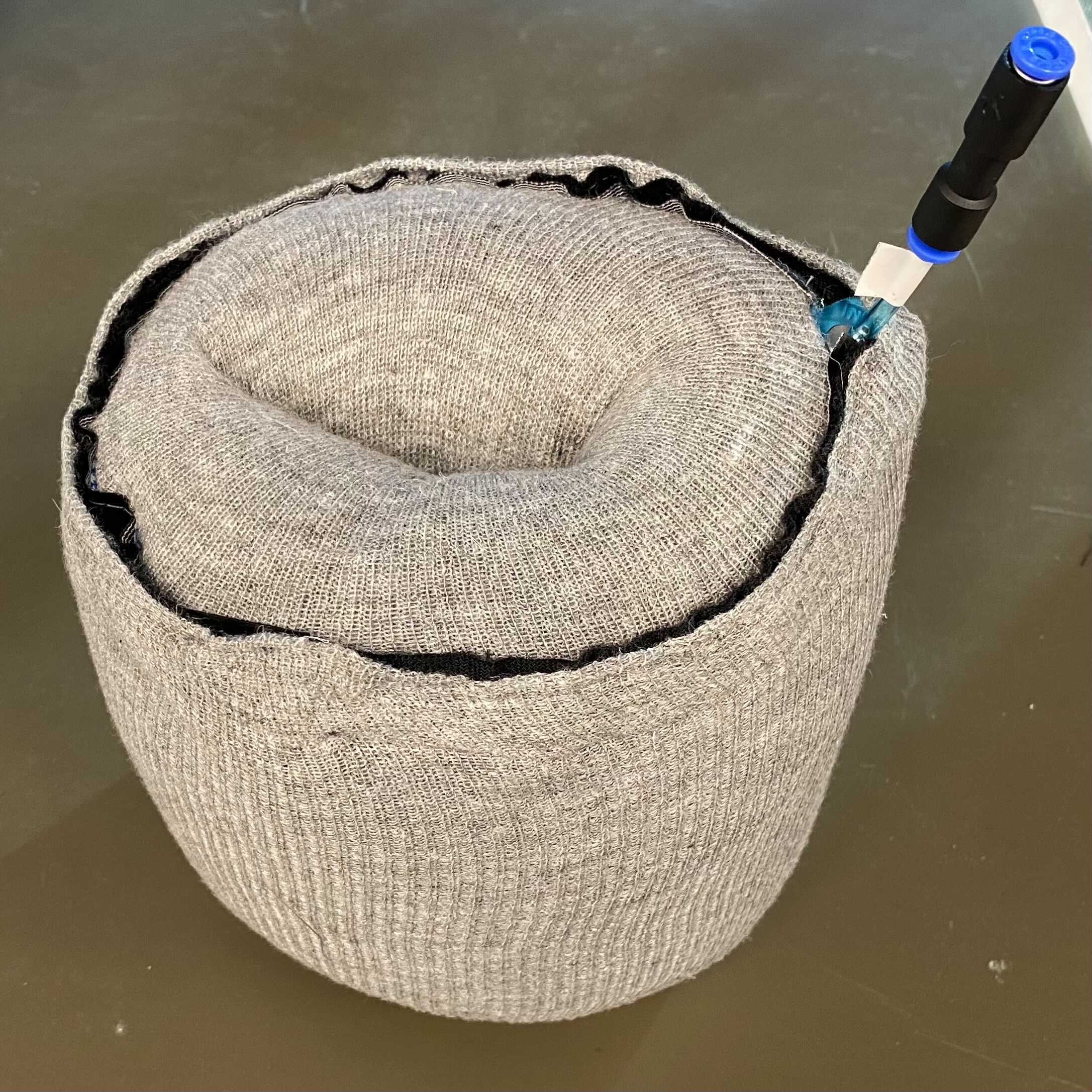}
  \caption{}
  \label{fig:floatie:covered}
\end{subfigure}\hfil % <-- added
\caption{Pressure-sensing chamber assembly: (a) Inflated armband with reduced port, (b) Partially opened Dyneema® fabric cover, and (c) robot-ready pressure-sensing chamber. }
\label{fig:floaties}
\end{figure}

Each arm is outfitted with four large, compliant, single chamber, air-filled pressure sensors, hereafter referred to as \emph{pressure-sensing chambers}, that cover each of the main links of the robot arm. The pressure-sensing chambers consist of off-the-shelf inflatable armbands equipped with barbed tube reducers (McMaster 3913N16) in the inflation ports (Fig.~\ref{fig:floatie:plain}) with tubes connecting each to central pressure sensing electronics.
Inflated armbands were chosen as the basis for our pressure-sensing chambers based on their passively stable performance during the kinematic exploration (Sec.~\ref{sec:mechanical}), availability, and ease of pressure sensing augmentation. The pressure-sensing chambers are held in place on the robot arm by high-friction, compliant foam structures which both prevent rotation around the link and form channels for wires and air tubes to run underneath the chambers (Fig.~\ref{fig:arm-tube-routing}).

The pressure-sensing chambers are covered in cut-resistant Dyneema® fabric (McMaster 7847T13) (Fig.~\ref{fig:floatie:covered}), which protects 
from sharp features that may be present in uncontrolled, cluttered environments.
%The system also eases collision avoidance requirements as accidental contact with the environment or clutter does not damage the robot or the environment.
%Collision avoidance requirements are also eased by this system because if the robot makes accidental contact with its environment there is a reduced risk of damage to both the environment and itself.
% Collision avoidance requirements are also eased by this system because if the robot makes contact, whether intentional or accidental, the softness and ability to react reduce the risk of damage to the robot and environment
Each cut-proof cover is sealed using hook and loop and can be opened for %armband 
repair and servicing, as well as for installing additional sensing layers~(Fig.~\ref{fig:floatie:halfcover}).
Each pressure-sensing chamber provides a scalar pressure value from a high accuracy Honeywell MPRLS0025PA00001A sensor centrally located at each arm's elbow.

\subsection{Module ii: \Softbubble ``Paw'' End Effectors} 

\begin{figure}[t]
    \centering % <-- added
    %\begin{subfigure}{0.45\textwidth}
    \includegraphics[width=0.50\columnwidth]{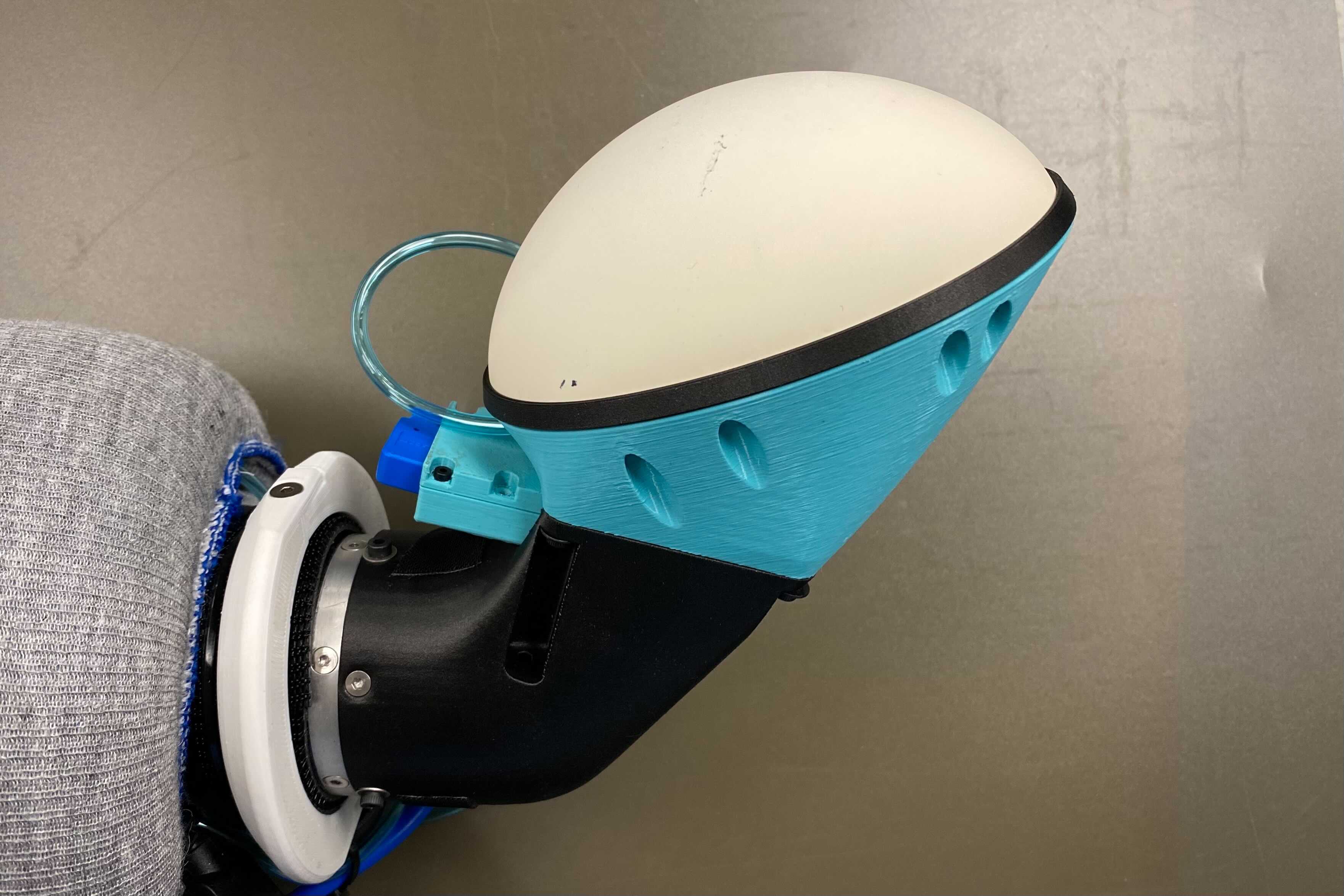}
%    \caption{\Softbubble ``paw'' module end effectors on our system.}
    \caption{Our \Softbubble ``paw'' module end effectors.}
    \label{fig:bubble-paw}
\end{figure}

The bimanual system includes two tactile sensing ``paws'' as end effectors that add no additional degrees of freedom (Fig.~\ref{fig:bubble-paw}). The compliant paws use \softbubble gripper modules~\cite{Kuppuswamy2020} and are able to sense internal pressure, dense contact geometry depth images, contact patches, and shear forces. The paws are used in both whole-body enveloping grasps 
%like those shown in 
(Fig.~\ref{fig:punyo-bimanual-grasp}) as well as palm-only grasps.

\subsection{Module iii: Force/Geometry Sensors on Chest}

\begin{figure}[t!]
    \centering % <-- added
\begin{subfigure}{0.23\textwidth}
  \includegraphics[width=\linewidth]{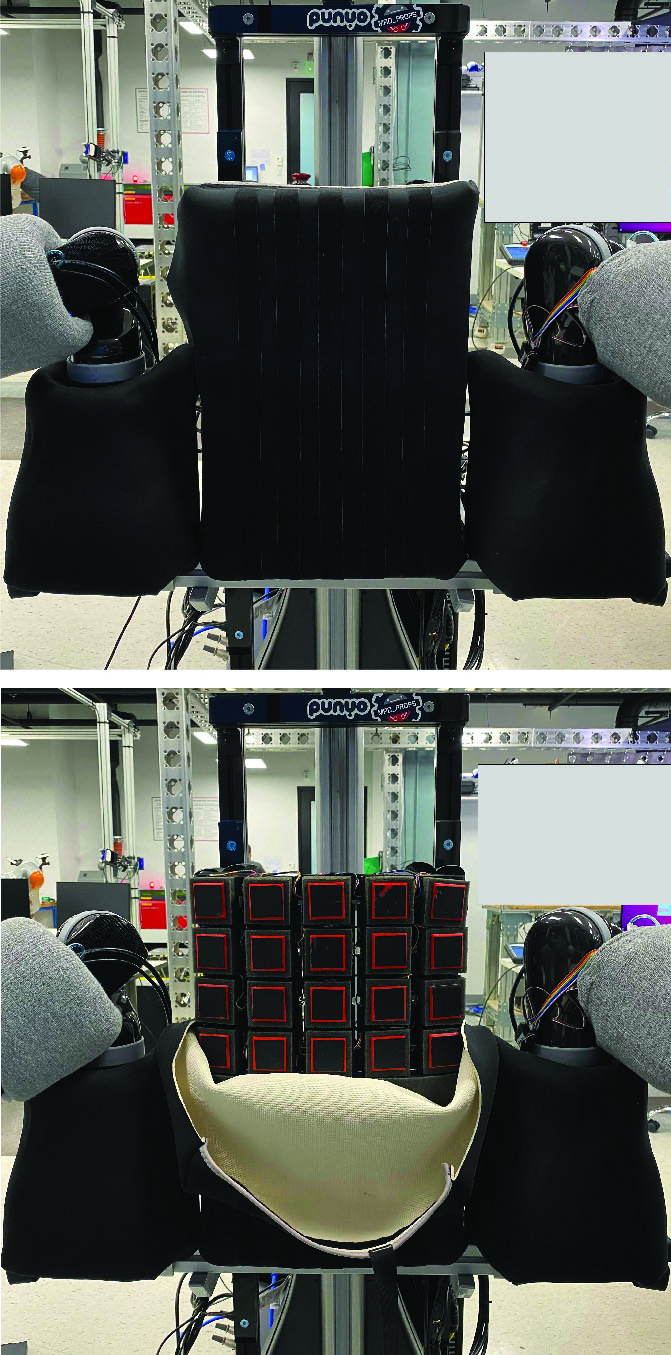}
  \caption{}
  \label{fig:inductive_chest_cover}
\end{subfigure}\hfil % <-- added
\begin{subfigure}{0.16\textwidth}
  \includegraphics[width=\linewidth]{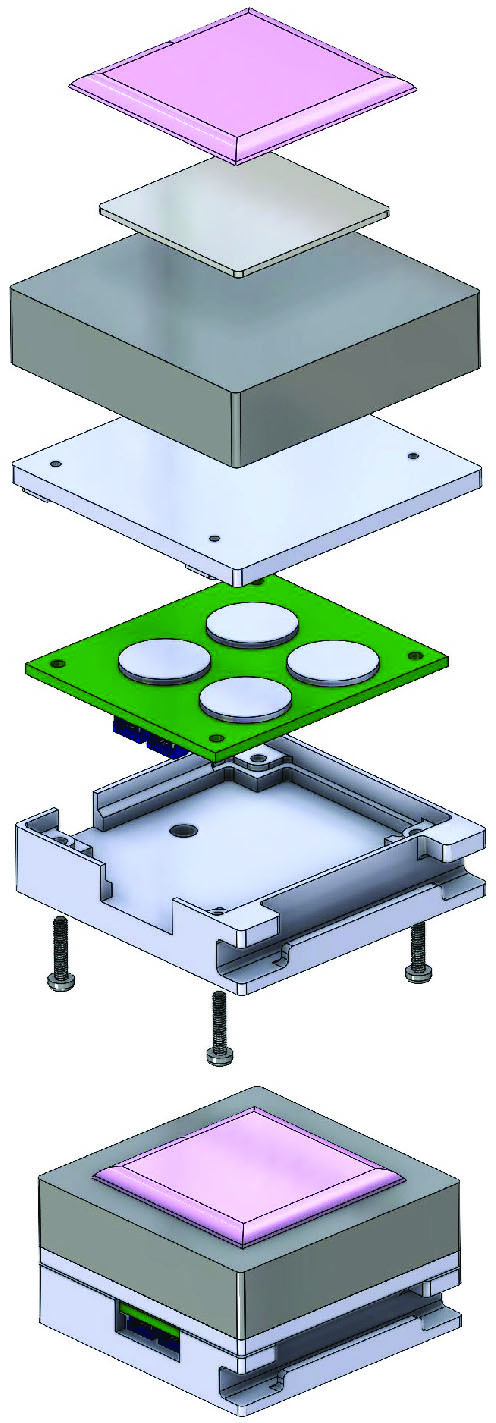}
    \caption{}
    \label{fig:inductive_chest_CAD}
\end{subfigure}\hfil % <-- added
%\caption{Force/geometry chest array: protective neoprene surface (a) covered and uncovered. (b) individual \forcegeometrysensor module exploded view and assembled.}
\caption{Force/geometry chest array: protective neoprene surface (a) covered and uncovered. (b) single \forcegeometrysensor module exploded view (\emph{top}) and assembled (\emph{bottom}).}
%\vspace{-3pt}
\label{fig:inductive_chest}
\end{figure}

The soft, sensorized chest of our upper-body system consists of a 4$\times$5 array of custom designed \emph{\forcegeometrysensor modules } (Fig.~\ref{fig:inductive_chest}). 
 
While our pressure-sensing chambers provide a single scalar pressure output, an individual \forcegeometrysensor module produces a directional force vector estimate during contact. 
%This measurement is done with a sensory board containing four coils that induces an electromagnetic field and measures the eddy currents introduced and varied by the conductive target in the closed system (in this design, the target is a metal plate). 
A sensory board containing four coils induces an electromagnetic field and measures the eddy currents induced in the conductive target (a metal plate) in the closed sensor system. The eddy currents are mapped to the distance and pose of the target, and, due to 
the foam separation layer, can be mapped to an implied force that created this displacement of the conductive target. With these sensors patterned across the contact surface of the robot chest, we are able to sense contact as the modules' sensing surface conform to the outer surface of an object.

%% KATE MOVED SO THAT IT MAKES MORE SENSE WITH THE TEXT

Mechanically, the assembly is as follows (Fig.~\ref{fig:inductive_chest_CAD}): Each \forcegeometrysensor houses a board with four coils and accompanying electronics in a 5~cm square by 13.5~cm tall ABS case, which mounts to the robot's chest frame. A 12.7~mm thick piece of soft polyurethane foam is glued to this ABS case, and on top of that there is a thin metal plate sandwiching the foam between itself and the ABS case. Unlike the pressure-sensing chambers, this design does not rely on inflated structures to achieve its softness. This allows for many types of compressible materials to replace the current foam, such as anisotropic 3D printed lattices, to obtain different properties and force vs. displacement profiles. This design allows us to tune the softness and sensing of the \forcegeometrysensor modular chest structure.

The array of \forcegeometrysensors  sits on a reconfigurable chest frame structure composed of two crossbars with five vertical slats where the sensor modules are mounted. The crossbars can be changed out to modify the curve of the chest; its current configuration has a slight convex curve with the surface of each slat (and thus the base of each \forcegeometrysensor) angled 170$^\circ$ from those horizontally adjacent. This curve could be modified to have a more or less extreme angle between the slats, or to even be concave, in order to support different types of grasps.

The chest is outfitted in a neoprene cover striped in 3M high-friction tape (3M TB641) (Fig.~\ref{fig:inductive_chest_cover}). 
The neoprene material creates a smooth, uniform surface over the entire \forcegeometrysensor array, providing a protective layer for the sensors and bridging gaps between the modules. 
Meanwhile, the 3M tape provides a high friction surface for objects to interface with during manipulation, reducing the load carried by the arms during a grasp.

\section{Grasp planning and control}
\label{sec:planning_and_control}

% \naveen{Naveen will simplify, describing the controller in prose for mechanical-minded dummies like Alex}

\begin{figure}[t]
    \centering % <-- added
    %\begin{subfigure}{0.45\textwidth}
    \includegraphics[trim={0 20px 0 40px},clip, width=0.65\columnwidth]{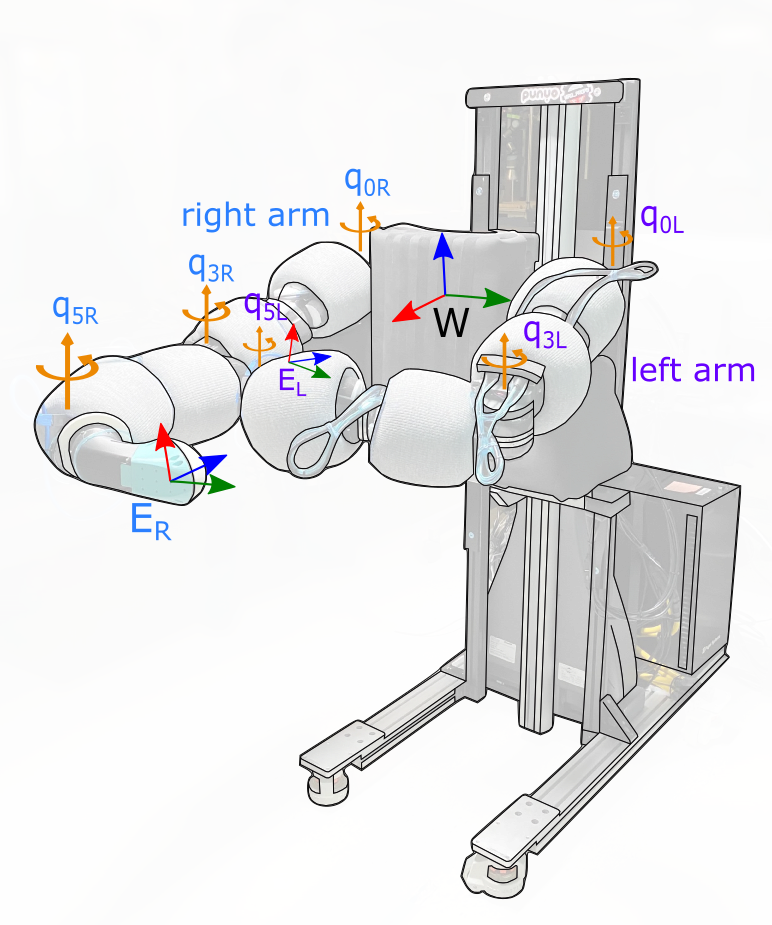}
    % \caption{Modeling the Punyo-1 robot: depiction of the frames and joints relevant to planar whole-body grasping used in Algorithm~\ref{alg:1grasping_primitive}.
    \caption{Punyo-1 model: depiction of the frames and joints relevant to planar whole-body grasping (Algorithm~\ref{alg:1grasping_primitive}).
    \naveen{reference in text}}
    \label{fig:punyo_frames}
\end{figure}

Exploiting mechanical intelligence often leads to simplifications in the control strategy for intelligent systems. In our case, taking advantage of our compliant hardware enables simple tactile feedback-based control schemes for grasping large objects. The tactile feedback is comprised of the sensed pressure at each of the pressure-based sensors, specifically the pressure-sensing chambers on the arms and \Softbubble paws on Punyo-1.
%; see Fig.~\ref{fig:punyo_frames}. 
For the scope of this paper, we make an assumption that the change in sensed pressure post-contact is strictly positively correlated to external contact forces, and therefore developed a grasp primitive based on a switching controller that is regulated by pressure changes. The Kinova Jaco SDK natively provides an interface for commanding joint velocities, and thus, our grasp primitive is designed in velocity space.

\begin{algorithm}[thpb]
\caption{Whole-body grasping primitive}
\label{alg:1grasping_primitive}
\begin{algorithmic}
  \State $P = [P_0, P_1, \ldots P_5 ]$\; \textit{\color{red}\# Per link relative pressure readings}\;
  \State $q_d = q_{d_0}$ \textit{\color{red}\# Initial condition of arm}\; 
  \State $P_G = [P_{G_0} \ldots P_{G_5}]$\;  \textit{\color{red}\# Grasp pressure change threshold}\;
  \State $j_S = [0, 3, 5]$  \textit{\color{red}\# Joints involved in the grasp}\;
  \State $i = 0$  \textit{\color{red}\# Current state of policy}\;
  \State $\dot{q}_G = [\dot{q}_{G_0}, \dot{q}_{G_3}, \dot{q}_{G_5}]$  \textit{\color{red}\# Desired joint velocities for each grasp stage}
    
\While{$i<3$}:
{
  \State \textbf{for each} $arm \in [left, right]$\;  \textit{\color{red}\# Repeat for each arm}
  {
    \hspace*{10mm} \State \textbf{command} $\dot{q}_G(i)$ on joint $j(i)$ and $0$ on others   \textit{\color{red}\# Move $1$ joint at a time} \;
  }
  \If{\textit{any}($P > P_G$)}
        \State $i = i+1$ \; \textit{\color{red}\#Transition to next stage of grasp}
  \EndIf
}
\EndWhile
% \\\textbf{Return}
\end{algorithmic}
\end{algorithm}

Using the grasp primitive defined in Algorithm~\ref{alg:1grasping_primitive}, we command each of the 7DOF arms independently while a low-level controller monitors for self-collisions and feasible motions at 100 Hz.
Fig. \ref{fig:punyo_frames} depicts the joints relevant to both the primitives and the world, as well as the end effector reference frames. 
%%% KATE FINDS THIS SENTENCE CONFUSING; TODO
Our grasps are designed primarily to activate joints $0$, $3$, and $5$ (or shoulder, elbow, and wrist) in that order. 

An example motion policy for whole-arm grasping for the right arm is defined by first deciding a convenient pre-grasp configuration $q(t_0)$; for instance, the arm oriented to be planar and curved inwards towards the target object with the soft face of the bubble palm facing towards the expected contact direction. All of the results presented in this paper used the same desired velocities, which for each stage of the grasp (denoted by $i$ in Algorithm \ref{alg:1grasping_primitive}), is given by
$\dot{q}_G = [-0.5, -0.25, -0.25]$. The terminal pressure threshold starts at ${P_G} = 20[1, 1, 1, 1]^T hPa$ - this magnitude corresponds to a gentle contact useful for human-robot interactions (e.g., hugging, Fig.~\ref{fig:punyo-hug}), though we used higher thresholds to achieve more stable grasps of larger objects. 

\section{Empirical testing and observations}
\label{sec:results}
Our hypothesis is that softness leads to improvements in whole-body control and interaction with objects.
We tested the Punyo-1 hardware through grasping tasks on a variety of domestic objects including a laundry hamper, bins, a water jug, stock pots, and a fish tank. In each case, we were able to generate contact-rich, stable grasps %. This was all done 
without external perception and without prior models of manipuland shape. 
%In this section, we present the analysis of the performance of the system.

\subsection{Softness and whole-arm grasping}
\label{sec:soft_v_rigid}

%%% KATE HAD TOO MANY QUESTIONS HERE AND REWORD SOME OF THE BELOW PARAGRAPH
%In order to test our hypothesis that softness leads to improvements in whole-body control and interaction with objects, we compared the performance of a hard surface Jaco arm (hard robot condition) against a soft Punyo-1 arm (soft robot condition) by grasping a series of cylindrical, weighted pots. We ran 3 trials of grasps for 4 increasing pot sizes (11.4 cm, 17.2 cm, 22.9 cm, and 31.1 cm) on both a single arm hard robot (Fig. \ref{fig:comp-grasp-test-hard}) and soft robot (Fig. \ref{fig:comp-grasp-test-soft}). 

\begin{figure}[t]
    \centering % <-- added
\begin{subfigure}{0.48\textwidth}
  \includegraphics[width=\linewidth]{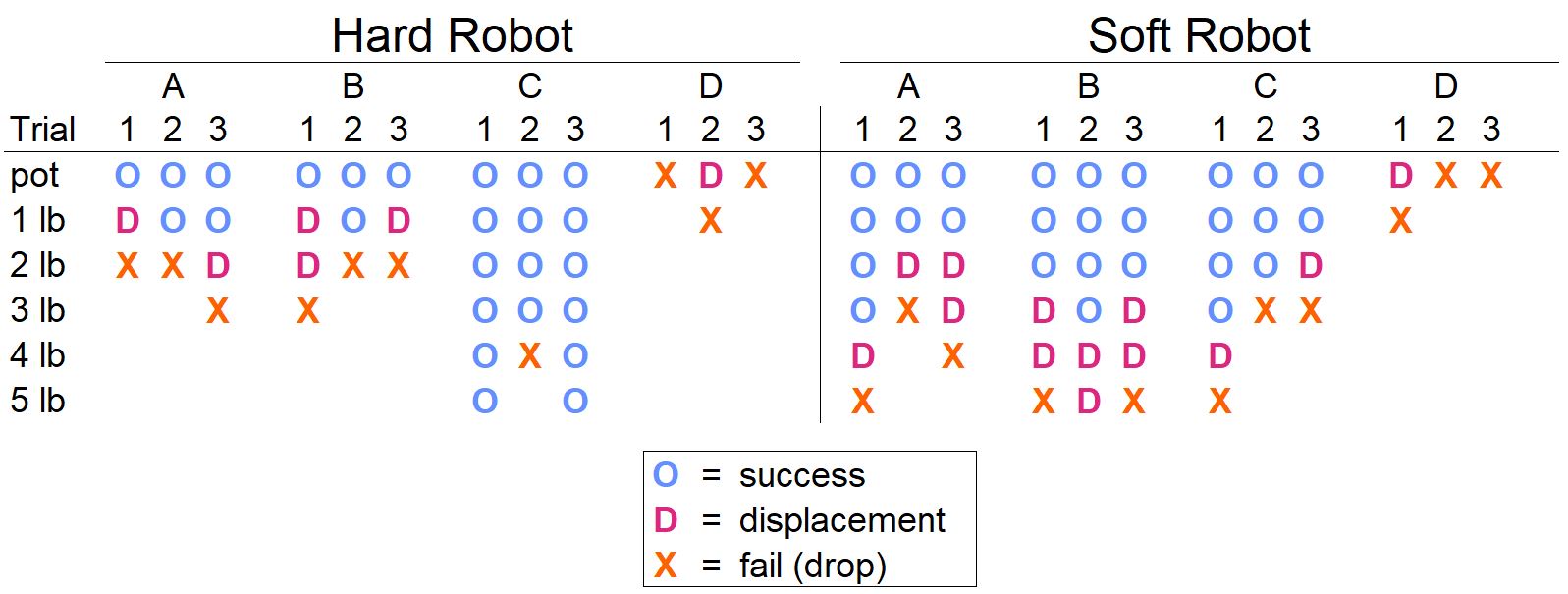}
    \caption{}
  \label{fig:comp-grasp-test-results}
\end{subfigure}\hfil % <-- added
\begin{subfigure}{0.23\textwidth}
  \includegraphics[width=\linewidth]{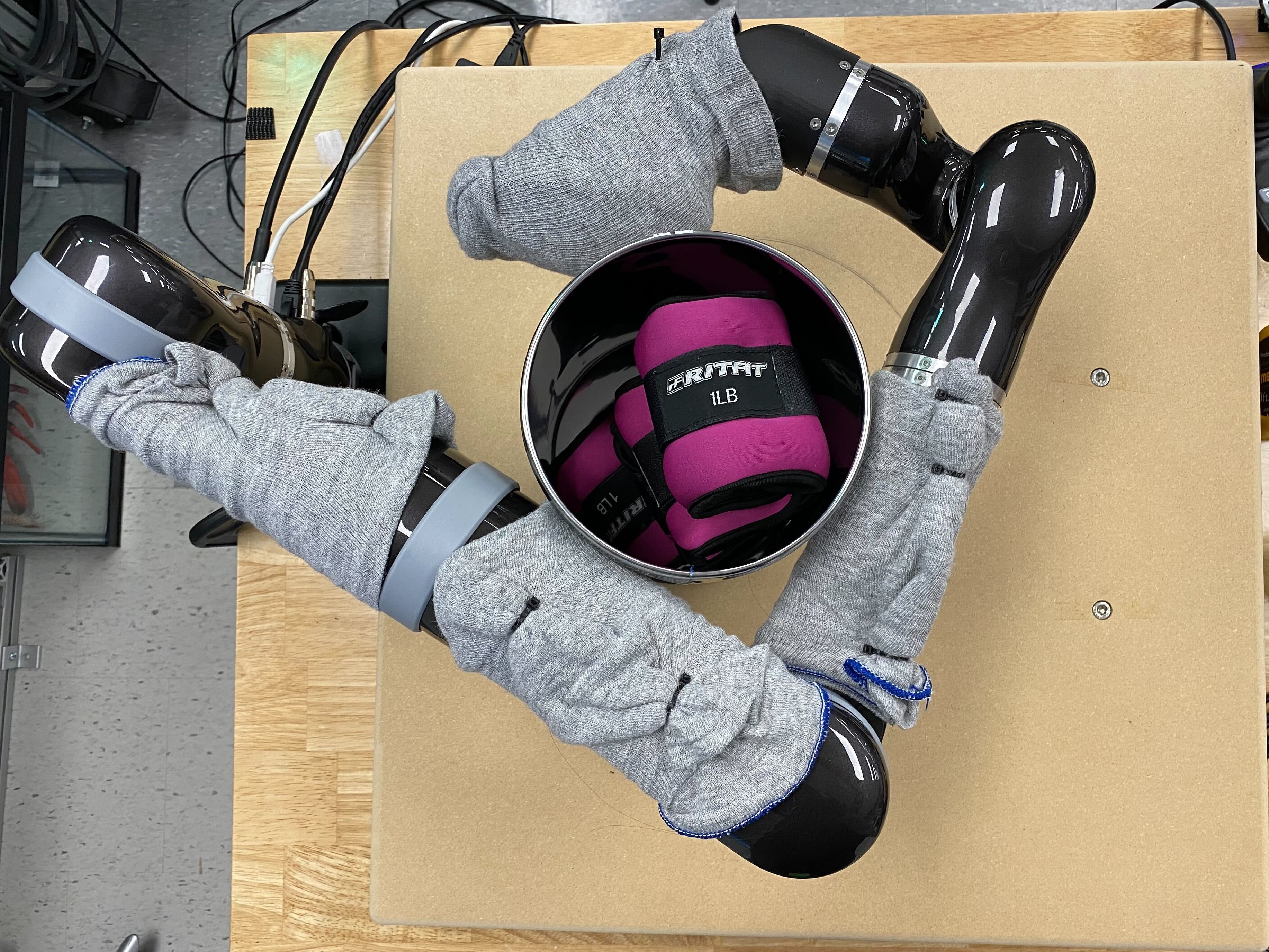}
  \caption{}
  \label{fig:comp-grasp-test-hard}
\end{subfigure}\hfil % <-- added
\begin{subfigure}{0.23\textwidth}
  \includegraphics[width=\linewidth]{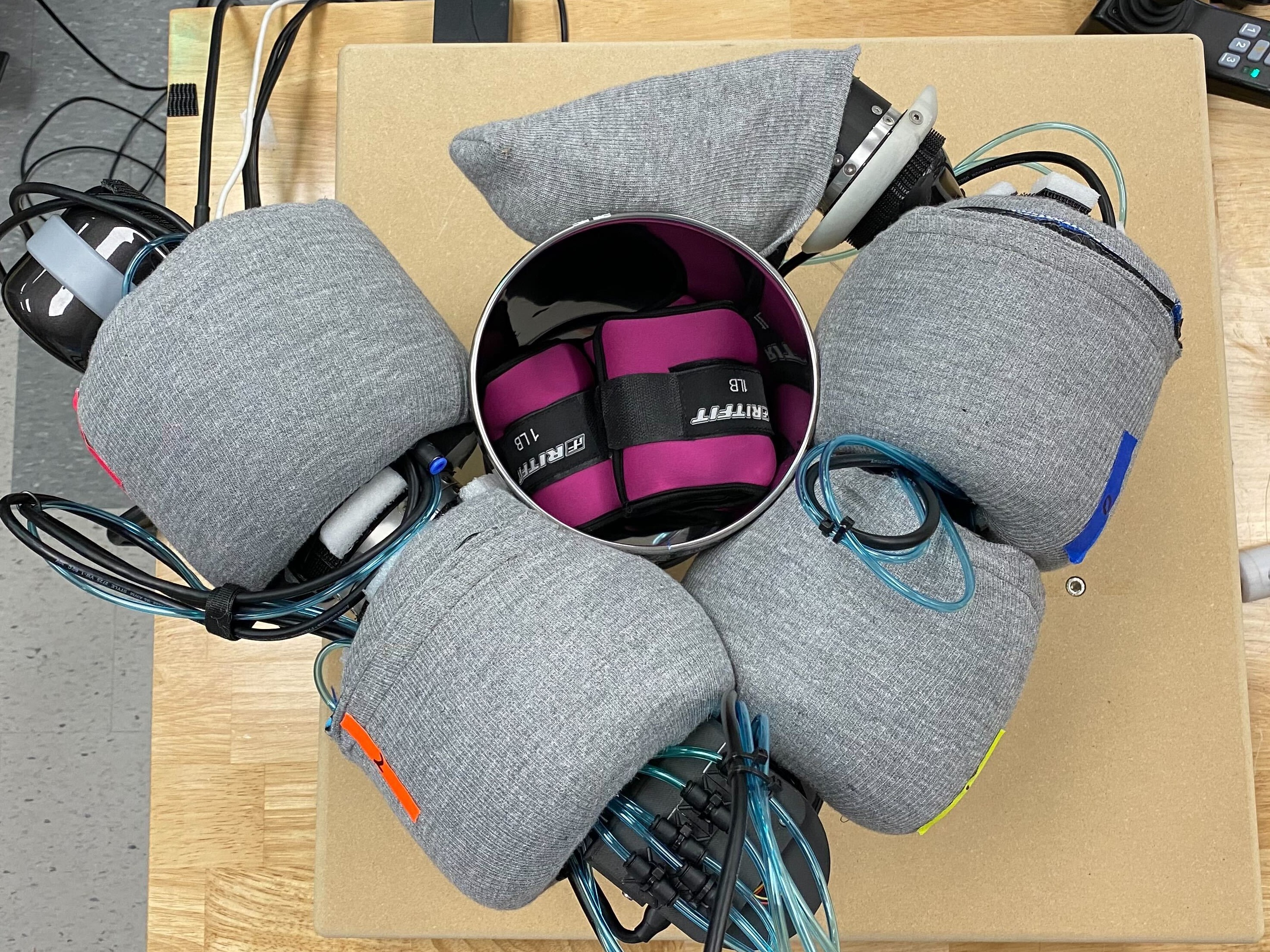}
  \caption{}
  \label{fig:comp-grasp-test-soft}
\end{subfigure}\hfil % <-- added
\caption{Results from a comparative study between a hard robot (a Jaco arm covered in Dyneema® fabric) and a soft robot (a Punyo-1 arm) grasping a series of a weighted cylindrical pots. (a) Test results from 3 trials of pot grasps for 4 increasing pot diameters, where the pot diameters  are as follows: A = 11.4 cm, B = 17.2 cm, C = 22.9 cm, D = 31.1 cm, and empty pot weights are as follows: A = 0.9 lb, B = 0.7 lb, C = 2.6 lb, D = 4.3 lb, (b) Hard robot grasping Pot B, (c) Soft robot grasping Pot B.}
\label{fig:softness-v-unmodified}
\end{figure}

In our first test, we focus on \emph{whole-arm grasping}; we compared the performance of the same Punyo-1 arm with 2 conditions (\emph{hard, soft}) by grasping a series of cylindrical, weighted pots. 
Fig. \ref{fig:comp-grasp-test-hard} shows the hard robot condition, and Fig. \ref{fig:comp-grasp-test-soft} shows the soft robot condition.
For consistency between the conditions, we minimized the difference in surface friction by outfitting the hard robot in the same Dyneema® material as the soft robot. 

We tested the hard robot condition vs. the soft robot condition over 3 trials of grasps for 4 increasing pot diameters: (11.4 cm, 17.2 cm, 22.9 cm, and 31.1 cm).
For each trial, a pot was placed on an identical location on a scissor lift table and the arm was teleoperated
\footnote{The same expert executed all trials: teleoperating the robot and identifying each grasp.}  
to secure the pot in a whole-arm grasp until the torque-based threshold was triggered via a visual LED indicator on the robot’s joystick. 
%Note that the same expert executed all trials: both teleoperating the robot and identifying each grasp's completion.
Once a grasp was visually identified, the table was lowered and weights were added to the pot in 1 lb increments %until either 5 lb was reached 
up to 5 lbs
or until the grasp failed. There were 3 potential outcomes: 1) \emph{displacement} if the pot shifted in the grasp while the table was lowered or weights were added but settled again, 2) \emph{failure} if the base of the pot landed back on the table, or 3) \emph{success} otherwise. %Grasp displacement was noted when the object shifted in the grasp but settled again.

% Claim 1: in general, the soft robot is able to grasp a larger variance of object sizes while the hard robot is only super successful in one specific configuration
The results in Fig. \ref{fig:comp-grasp-test-results} show that
the hard robot is capable of maintaining a stable grasp only for a particular configuration and object size (pot C, 22.9 cm)% than the soft robot
. However, in general, the soft robot is able to grasp a larger variance of object sizes. Neither robot was able to successfully grasp the largest object (pot D, 31.1 cm).%, but initial pot weight should be considered in this circumstance (pot D, 4.3 lb).

% Claim 2: elastic v. plastic
%Talk about weight here (b/c that's where displacement comes in)
Under load, the soft robot fails more gracefully as load increases than the hard robot. During testing, when displacement was observed, the object settled into a stable configuration more often in the soft robot's case as the weight increased rather than failing outright (13 occurrences). However, in the hard robot's case, displacement instances were less frequent (5 occurrences) and the failure modes were abrupt. 
%We hypothesize moving forward that there is a chance for grasp recovery through exploiting tactile feedback during displacement.
For the smaller objects%(pot A = 11.14 cm, pot B = 17.2 cm)
, Fig. \ref{fig:comp-grasp-test-results} 
shows that the soft robot is able to handle the addition of more weight than the hard robot (4 lb and 1 lb, respectively).

% There are 2 key characteristics of softness that are visible in the plots. Firstly, it can be seen that the peak velocities at  

\subsection{Whole-body grasping}

% \naveen{We need this to be something as simple as: We tested the combination of our hardware and controller on a variety of objects, including the hamper, bins, pasta pot, a fish tank, and in general, we were able to create contact-rich, stable grasps. This was all done without external perception of manipuland shape and without a priori models of said objects. Here are some pictures of the grasp controller running. The grasp stability was qualitatively tested by lifting and perturbing most objects.

% We also never mention paw-only grasps in the paper, do we?

% As far as initial pose variation experiments go, I think it is much more fair to say that we didn't explicitly/precisely control initial pose of grasped objects and the hard-controller combo performed well regardless}

\begin{figure}[h!t]
    \centering % <-- added
\begin{subfigure}{0.23\textwidth}
  \includegraphics[width=\linewidth]{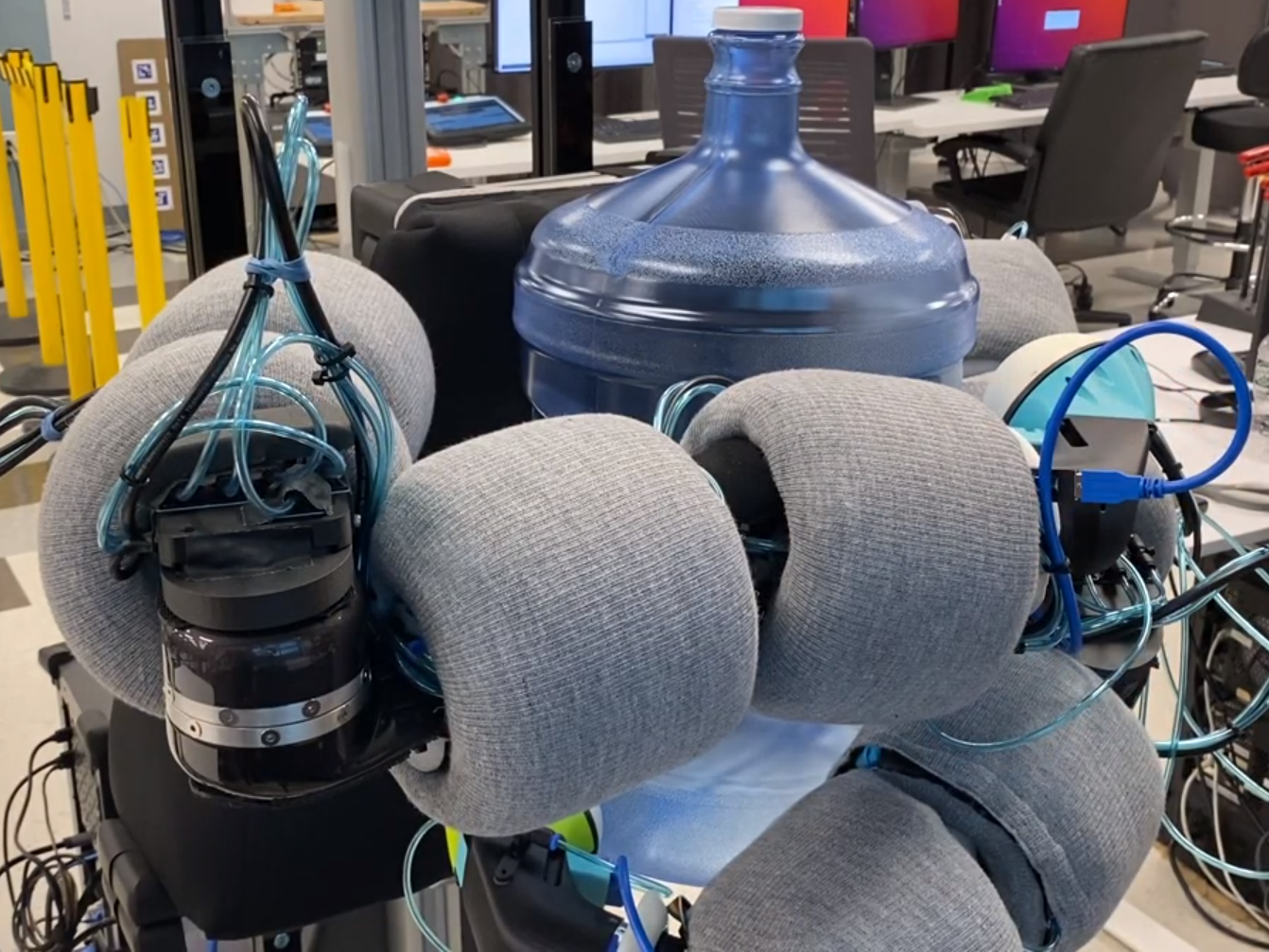}
  \caption{}
  \label{fig:waterbottle-real}
\end{subfigure}\hfil % <-- added
\begin{subfigure}{0.23\textwidth}
  \includegraphics[width=\linewidth]{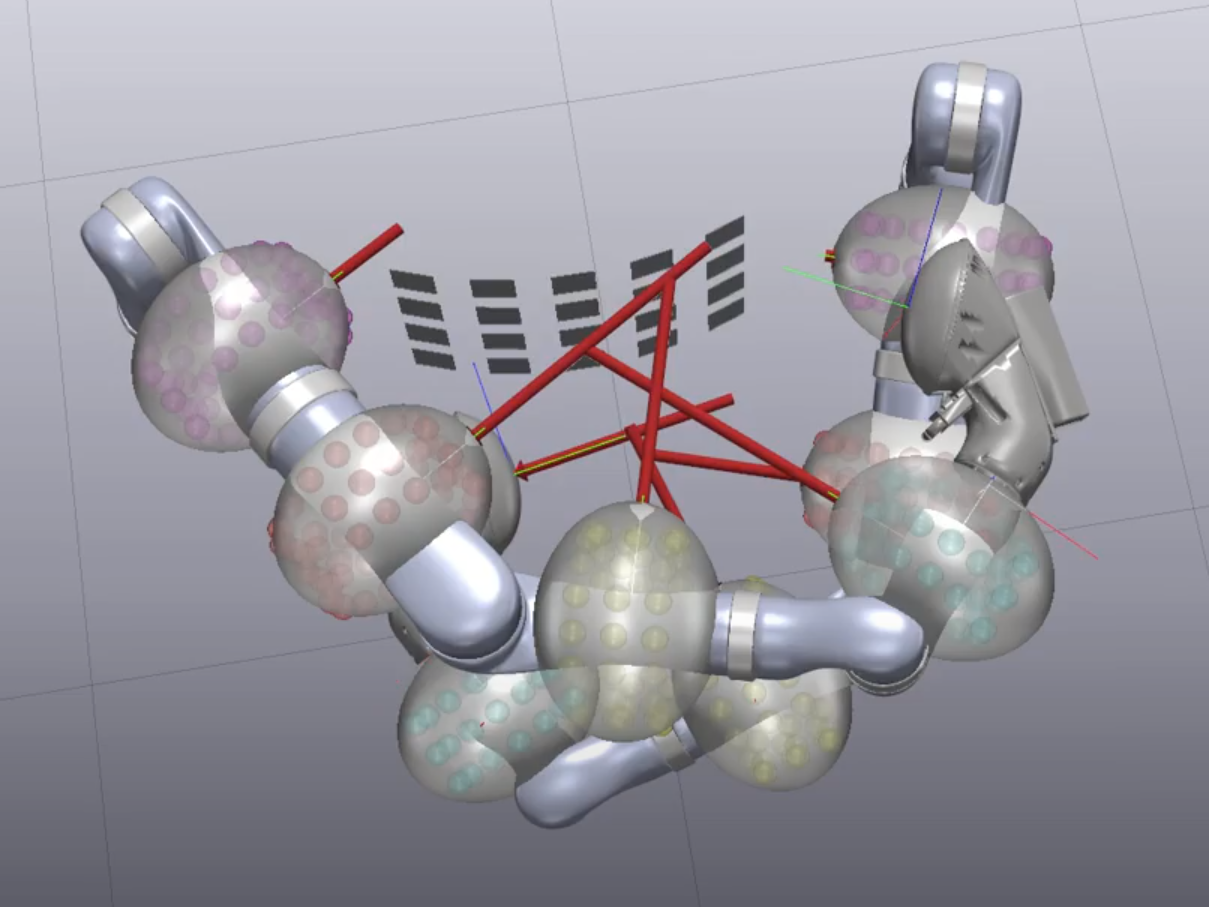}
  \caption{}
  \label{fig:waterbottle-vis}
\end{subfigure}\hfil % <-- added
\begin{subfigure}{0.23\textwidth}
  \includegraphics[width=\linewidth]{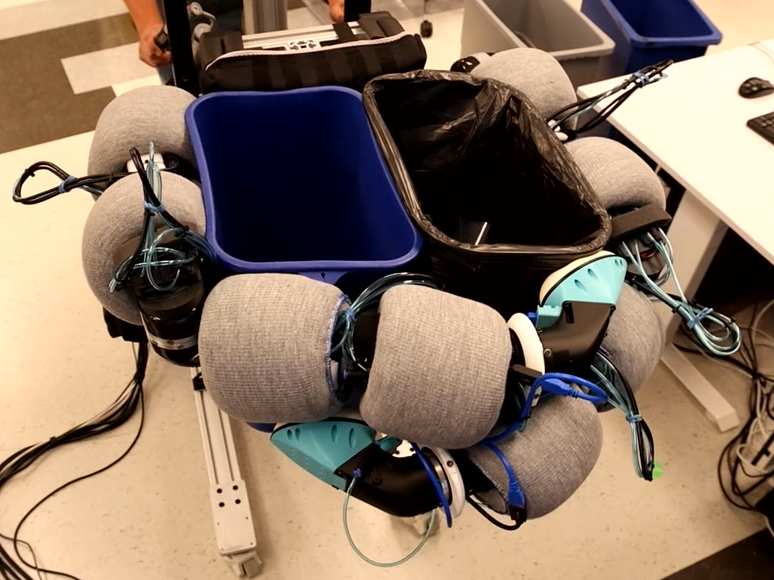}
    \caption{}
  \label{fig:dualtrashbin-grasp}
\end{subfigure}\hfil % <-- added
\begin{subfigure}{0.23\textwidth}
  \includegraphics[width=\linewidth]{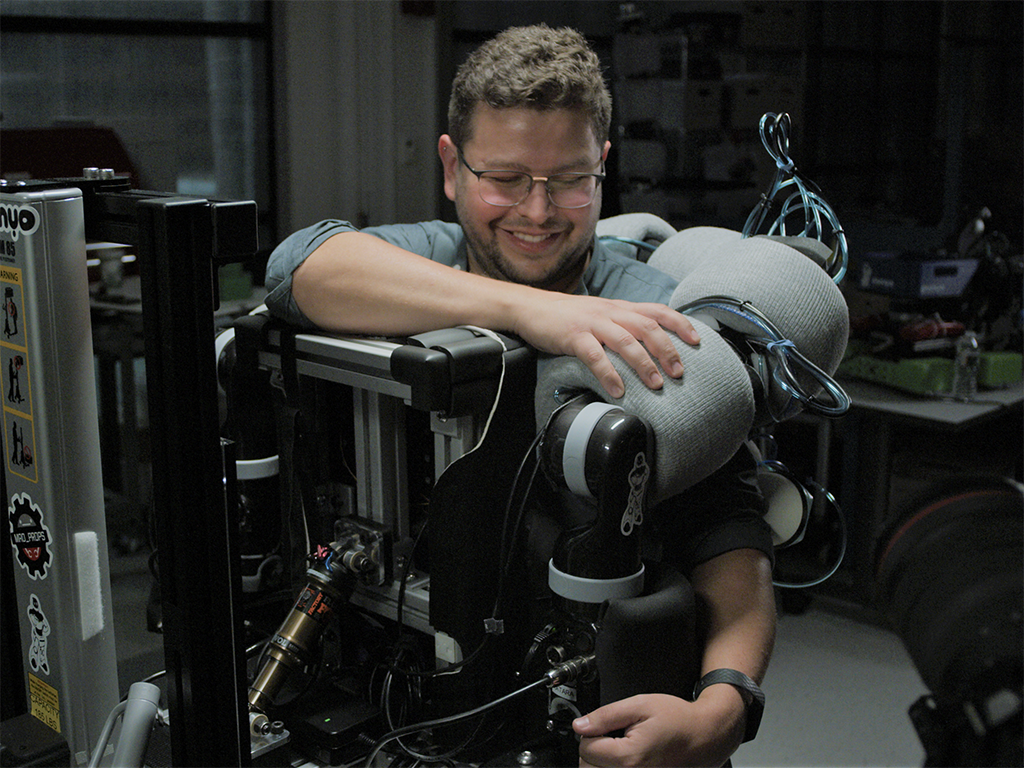}
    \caption{}
  \label{fig:punyo-hug}
\end{subfigure}\hfil % <-- added
\caption{Robot grasping examples: (a) stable water bottle grasp and (b) corresponding sensor visualization, (c) dual waste bin grasp, and (d) robot-human hugging.}
\label{fig:punyo-bimanual-grasp}
\end{figure}

In order to demonstrate the whole-body manipulation capabilities of Punyo-1, we tested and qualitatively analyzed the robustness of the resulting grasps on a set of objects. %(Fig.~\ref{fig:punyo-bimanual-grasp}). 
As previously noted, the grasp primitive (Algorithm~\ref{alg:1grasping_primitive}) does not have nor requires models of objects. 
We show that the system can handle more forceful tasks like carrying a large water bottle (Fig.~\ref{fig:waterbottle-real}) or grasping two bins simultaneously (Fig.~\ref{fig:dualtrashbin-grasp}). Initial testing shows that the upper-body can reliably achieve a stable grasp that is robust to antagonism. We can also use the same controller in a more delicate human-robot hugging interaction by lowering the grasp controller's pressure threshold (Fig.~\ref{fig:punyo-hug}). 

\subsection{Robustness of grasps to initial conditions}
\label{sec:exp_initial_conditions}
% Our platform and controller are able to handle a range of variability in initial object pose. 

\begin{figure}[!htbp]
\centering
\begin{subfigure}{0.23\textwidth}
  \centering
  \includegraphics[trim={500px 50px 550px 180px},clip,height=10em]{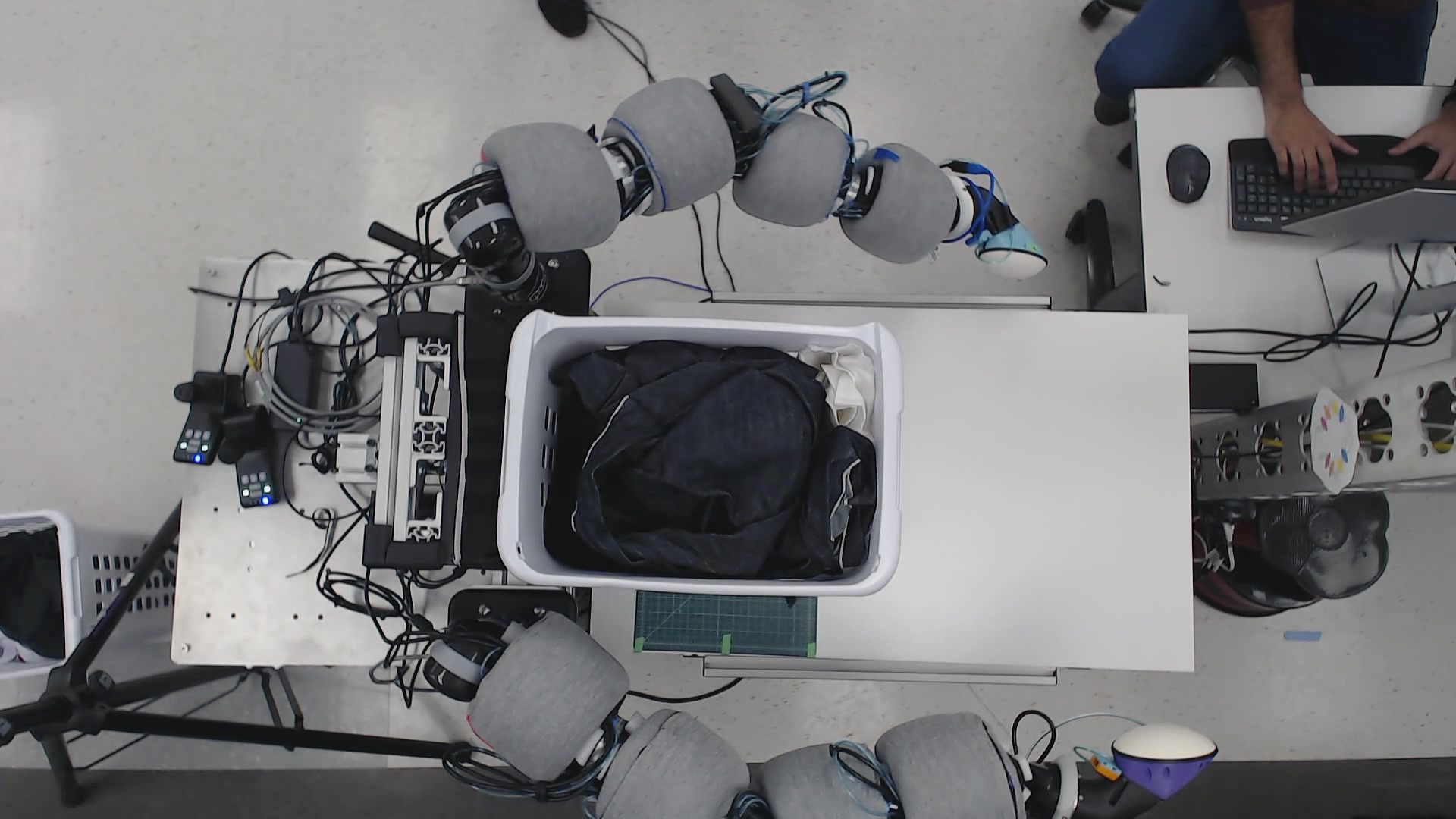}
  \caption{}
  \label{fig:hamper-pose-a-start}
\end{subfigure}\hfil % <-- added
\begin{subfigure}{0.23\textwidth}
  \centering
  \includegraphics[trim={500px 50px 550px 150px},clip,height=10em]{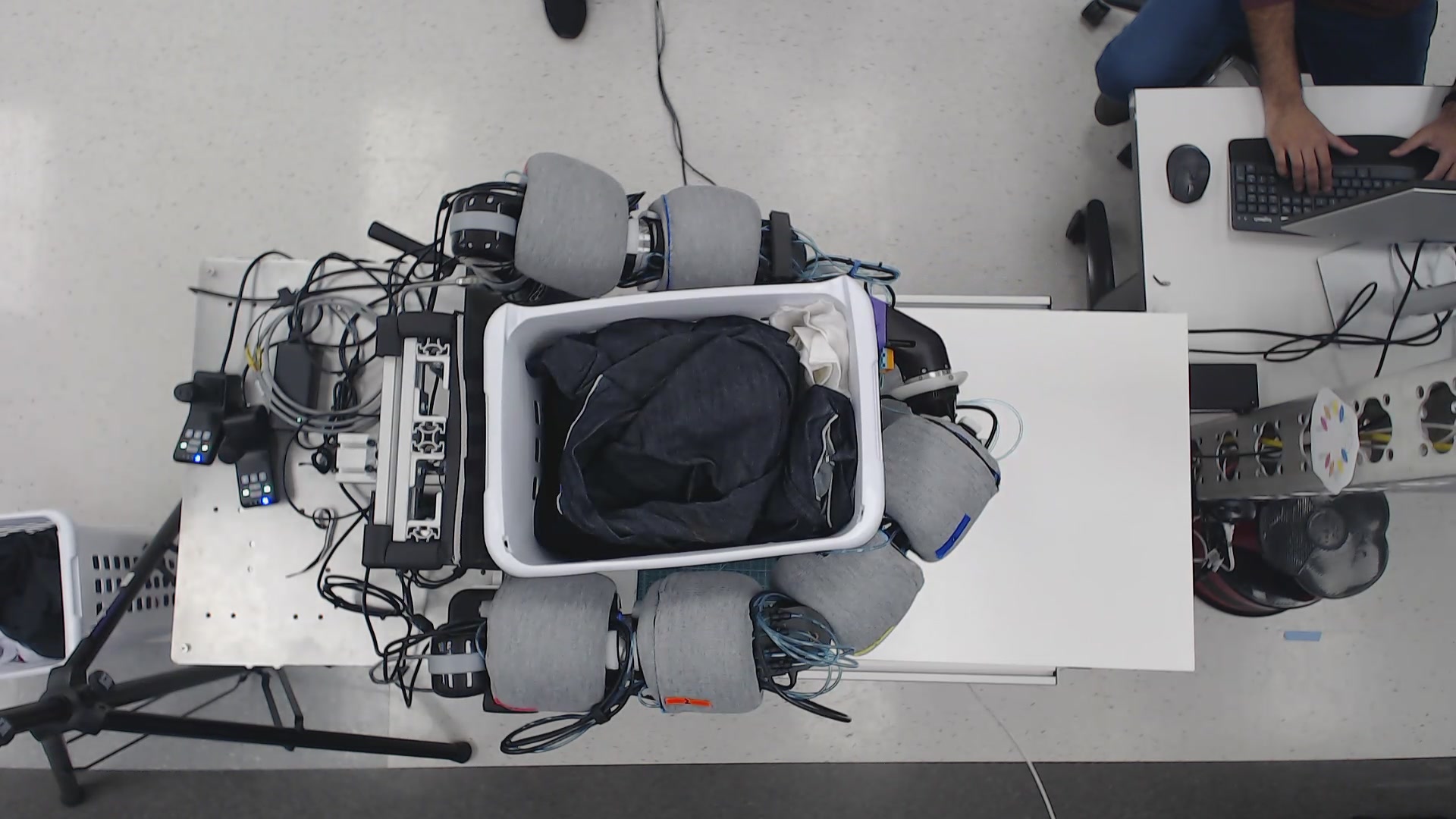}
  \caption{}
  \label{fig:hamper-pose-a-end}
\end{subfigure}\hfil % <-- added
\begin{subfigure}{0.23\textwidth}
  \centering
  \includegraphics[trim={500px 50px 550px 150px},clip,height=10em]{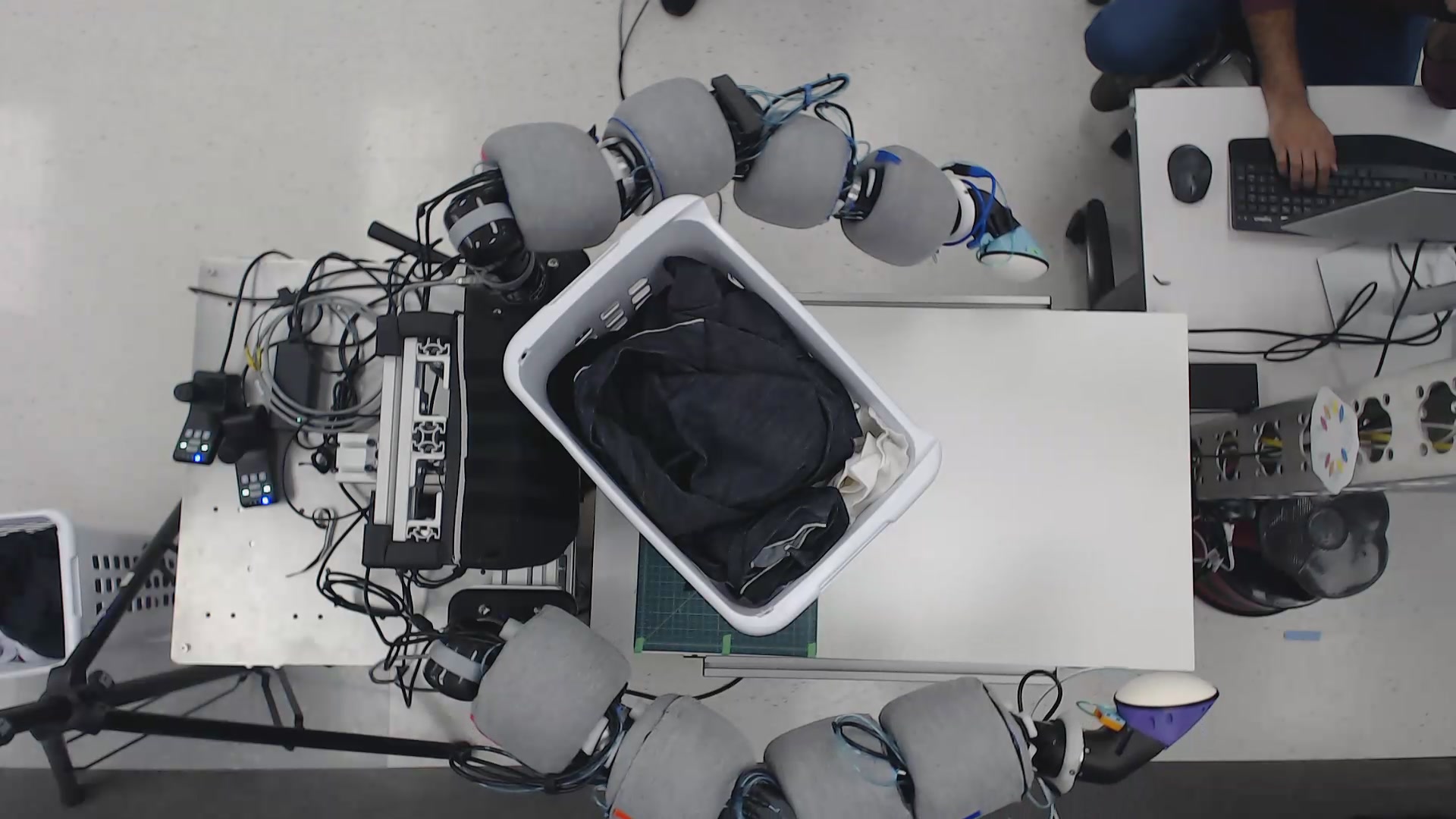}
    \caption{}
  \label{fig:hamper-pose-b-start}
\end{subfigure}\hfil % <-- added
\begin{subfigure}{0.23\textwidth}
  \centering
  \includegraphics[trim={500px 50px 550px 150px},clip,height=10em]{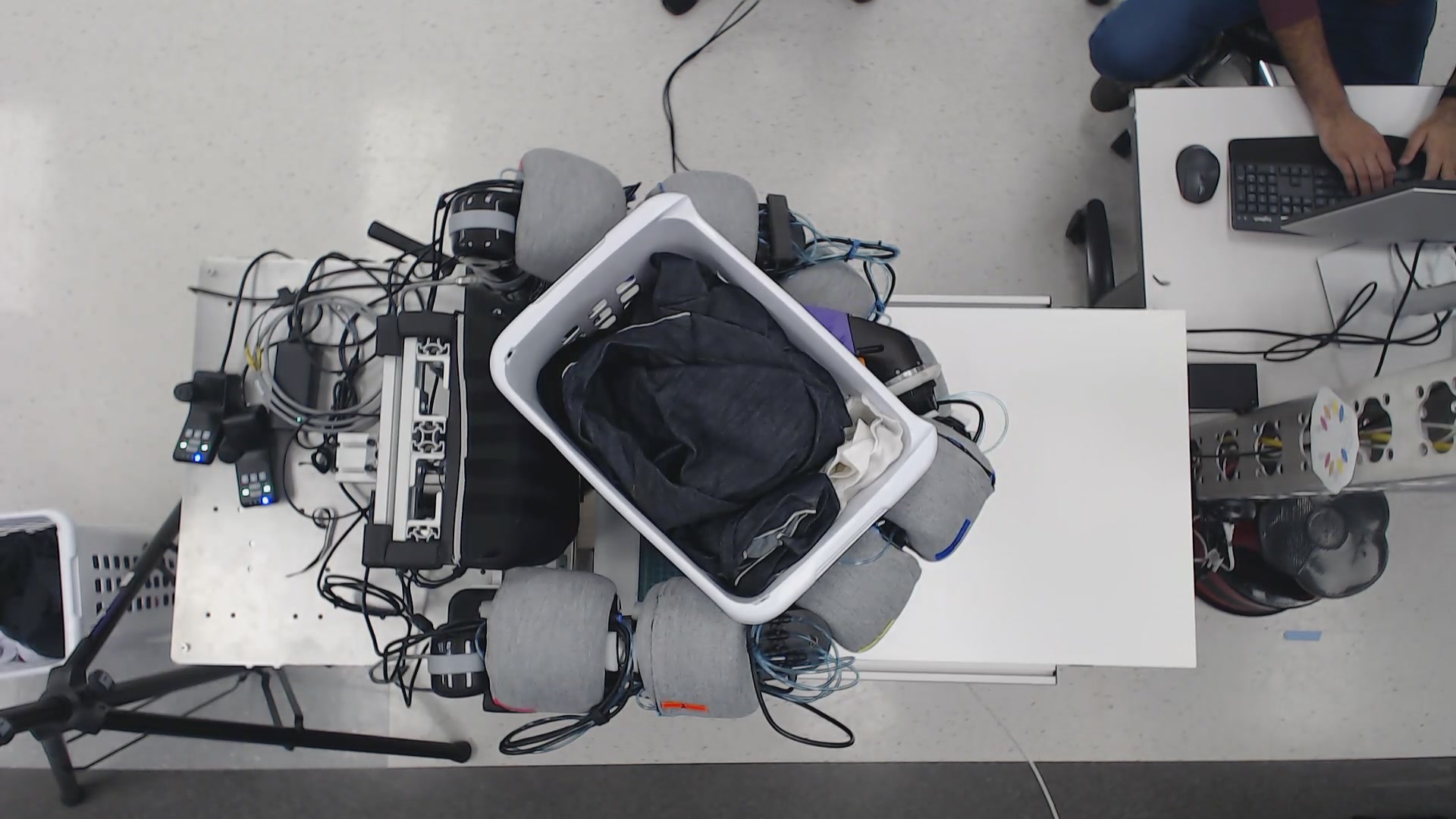}
  \caption{}
  \label{fig:hamper-pose-b-end}
\end{subfigure}\hfil % <-- added
\begin{subfigure}{0.40\textwidth}
    \centering
    \vspace{1em}
    \includegraphics[width=\linewidth]{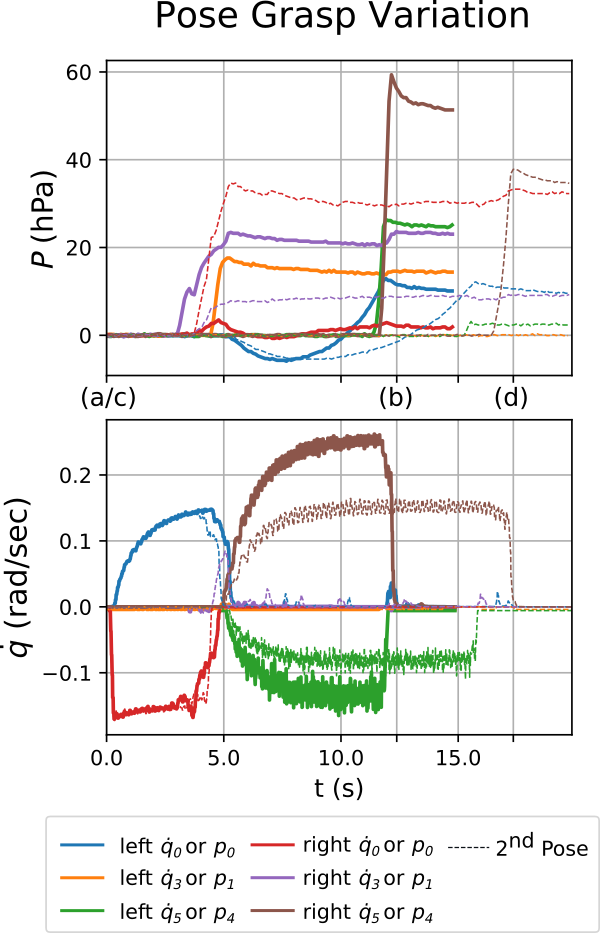}
    % \includesvg[width=\linewidth]{figures/hamper-time-series.svg}
    \caption{}
    \vspace{1em}
    \label{fig:hamper-grasp-timeseries}
\end{subfigure}\hfil % <-- added
\caption{Two grasp series with varied initial poses (a): $0^\circ$ and (c): $45^\circ$. The controller is unaware of the change in initial conditions reacting only to the tactile stimulus $P$. Note that in the $0 ^\circ$ trial (solid) the grasp completes faster (b) because sufficient pressure occurs early in the motion, while in the $45 ^\circ$ trial (dashed) the paw moves a further distance before meeting the threshold $P_{G}$ (d).  }
\label{fig:hamper-grasp-series}
\end{figure}

We explored the robustness of the grasp controller to varied initial conditions by changing the initial pose of the target manipuland (a laundry hamper) and examined the controller's input and output signals (Fig.~\ref{fig:hamper-grasp-series}). The controller had no knowledge that the initial pose changed from $0^\circ$ to $45^\circ$ (Figs.~\ref{fig:hamper-pose-a-start} and \ref{fig:hamper-pose-b-start}, respectively), yet still achieved a stable grasp (Figs.~\ref{fig:hamper-pose-a-end} and~\ref{fig:hamper-pose-b-end}) only using the tactile signal $P$ to control the velocities $\dot{q}$. 
% Our platform and controller are able to handle a range of variability in initial object pose. The adaptability and stability of our grasps is due to the highly compliant nature of the robot and controller.
The adaptability and stability of our grasps is due to the highly compliant nature of the robot.

\subsection{System durability}
\label{sec:durability}
Punyo-1's hardware proved durable when subjected to the repeated grasp testing detailed in this manuscript. 
The performance of the Dyneema® and neoprene fabric covers for the pressure-sensing chambers and \forcegeometrysensor chest array, respectively, provided confidence that the system could resiliently interact with objects without being punctured or imprinted. 
Collision avoidance requirements were also eased by the hardware because if the robot made contact, whether intentional or accidental, the softness and ability to react on sensor input reduced the risk of damage to the robot and environment.

%%%%%%%%%%%%%%%%%%%%%%%%%%%%%%%%%%%%%%%%%%%%%%%%%%%%%%%%%

\section{Discussion}
\label{sec:discussion}

It is evident from our grasping experiments that the addition of highly compliant surfaces to \emph{hard} robot manipulators and upper-body grasping surfaces enables stable whole-body grasps with a reduced reliance on object models, perception, planning, and control. The large, encompassing contact patches afforded by the surface deformation help to solve typical grasping problems mechanically, including generating ample friction and obtaining force and form closure. These results solidify our commitment to continued development of a modular soft, sensing hardware and algorithmic toolbox for augmenting hard robots and reducing the barrier to entry for robust, whole-body grasping. 
 
While our robot's compliance is playing a large role in grasp stability, the tactile sensor feedback available is currently underutilized. Air pressure signals from the arm's pressure-sensing chambers and end-effector paws are used to adapt each grasp to objects, but the chest's force/geometry, arm's surface contact position, and paw's shear and geometry signals are currently unused. To move from basic object grasping to force-based manipulation, it is important to further understand the roles these and other sensing modalities play in contact patch estimation, object state estimation, and force control.

Even at this early stage of understanding, the breadth of potential capabilities enabled by rich tactile sensing is apparent. The ability for a robot to sense contact, both intentional and incidental, on all surfaces is important in the domestic environment where space is tight, cluttered, unmodeled, and shared with humans. Robots with whole-body tactile sensing offer many opportunities for compelling physical human-robot interaction, collaboration, and communication.

\section{Future Work}
\label{sec:futurework}

%Our focus to date has been on leveraging the ``mechanical intelligence'' of ZYX. 
%% design philosophy of \emph{mechanical intelligence}, i.e., taking advantage of passive mechanical and material properties to solve problems~\cite{kim2013soft}
% Hardware

Expanding upon our development of soft, sensing modules for \emph{hard} robots, we will focus on techniques for localizing contact and estimating contact patches for sensors along the arms. 
We will continue to develop robot hardware agnostic mounting schemes and strategies for future modules. 
In addition, rigorous mechanical and grasp testing will lead to a better understanding of mechanical failures and module durability.

% Controls
As we make advances in tactile sensing hardware capabilities, we must also work on how to intelligently integrate the sensing into controls and planning. 
A more principled approach that uses both prior assumptions on object geometry and sensory feedback can enable dynamically interactive manipulation primitives.
These primitives can generate more insight into which sensing modalities are useful, guiding future hardware development.

Lastly, our hardware platform is well suited for physical human-robot interaction research. We will use this system to explore collaborative manipulation in the context of physical assistance in the domestic environment.

% Outro
We have only scratched the surface on integrating multimodal tactile sensing - Punyo-1 is only one possible module configuration. We will continue to use our hardware design philosophy as we design future modules, outfit more robots, and work towards robust robotic manipulation in the home.

%%%%%%%%%%%%%%%

% \subsection{Modules}

%%% 3 points to make:
% increasing mesh size of pressure sensors
% exploring layered sensor add-ons
% exploring pinch point modules

%Motivated by rigorous testing of our current system, we will be gaining continuous feedback for future module designs and overall system improvements.

%For example, this work explores and pushes the capabilities of coarse sensing by having single pressure chamber modules on each arm link. 
%Our next focus is increasing the sensor spatial resolution in order to deal with self-contact discrimination in extreme joint angles. 
%We are also investigating more bespoke fabrication techniques to better form-fit with robot surface geometries and to enable more layered sensing both over and underneath each module. 
%We plan to continue exploring more module designs in an effort to fully cover robot arms, including moving joints and pinch points.

%\subsection{Heavy Objects}

%%%%%%%%%%%%%%%
%Our current experimentation with whole-body soft sensing has focused on manipulating larger, unwieldy objects.
%Throughout the exploration and development of the soft, tactile sensing modules, we have learned that the underlying off-the-shelf robotic arms can be pushed beyond their advertised payloads to support heavier objects when utilizing whole-arm grasping techniques.
%These early experiments inspire future work in which we will optimize and quantify these capabilities of our system so that we may expand our object list to include items that are both large \emph{and} heavy. 

%%

\section*{Acknowledgements}
Thank you Sam Creasey, Tristan Whiting, Phoebe Horgan, Richard Denitto, Liam Rondon, Adeeb Abbas, and Sean Taylor for engineering, experimental, filming, and paper graphics support.

\balance

\bibliographystyle{IEEEtran}
\bibliography{IEEEabrv,reference} % Note the lack of whitespace between the commas and the next bib file. It won't work if there are any spaces.
\end{document}